\documentclass{article}

\PassOptionsToPackage{numbers, sort, compress}{natbib}

    \usepackage[preprint]{neurips_2024}

\usepackage[utf8]{inputenc} %
\usepackage[T1]{fontenc}    %
\usepackage{hyperref}       %
\usepackage{url}            %
\usepackage{booktabs}       %
\usepackage{amsfonts}       %
\usepackage{nicefrac}       %
\usepackage{microtype}      %
\usepackage{xcolor}         %
\usepackage{array}
\usepackage{microtype}
\usepackage{hyperref}
\usepackage{url}
\usepackage{booktabs}
\usepackage{times}
\usepackage{latexsym}
\usepackage{adjustbox}

\usepackage[compact]{titlesec}

\usepackage[T1]{fontenc}

\usepackage[utf8]{inputenc}

\usepackage{microtype}

\usepackage{inconsolata}

\usepackage{graphicx} %
\usepackage{graphicx} %
\usepackage{booktabs}
\usepackage{algorithm}
\usepackage{algorithmic}
\usepackage{graphicx}
\usepackage{array}
\usepackage{subcaption}
\usepackage{xcolor}
\usepackage{hyperref}
\usepackage{url}
\usepackage{xspace}

\newcommand{\RomanNumeralCaps}[1]{\MakeUppercase{\romannumeral #1}}

\newcommand{\contextual}{$a_c$\xspace}
\newcommand{\golden}{$a_g$\xspace}
\newcommand{\parametric}{$a_p$\xspace}
\newcommand{\question}{$q$\xspace}
\newcommand{\context}{$c$\xspace}
\newcommand{\score}{$s$\xspace}

\newcommand{\obQA}{open-book\xspace}
\newcommand{\cbQA}{closed-book\xspace}
\newcommand{\bn}{WACK\xspace}

  \usepackage{microtype}
\usepackage{hyperref}
\usepackage{url}
\usepackage{booktabs}
\usepackage{amsmath}
\usepackage{enumitem}
\usepackage{multirow}
\renewcommand\footnotemark{}

\title{Constructing Benchmarks and Interventions for Combating Hallucinations in LLMs}

\author{
Adi Simhi\textsuperscript{1} \hspace{1em} Jonathan Herzig\textsuperscript{2} \hspace{1em} Idan Szpektor\textsuperscript{2} \hspace{1em} Yonatan Belinkov\textsuperscript{1}\\
\thanks{\texttt{adi.simhi@campus.technion.ac.il},\texttt{\{jherzig,szpektor\}@google.com},\texttt{belinkov@technion.ac.il}} 
\textsuperscript{1}Technion -- Israel Institute of Technology\\
\  \textsuperscript{2}Google Research \\
}

\begin{document}

\maketitle

\begin{abstract}
Large language models (LLMs) are prone to hallucinations, which sparked a widespread effort to detect and prevent them. Recent work attempts to mitigate hallucinations by intervening in the model's generation, typically computing representative vectors of hallucinations vs.\ grounded generations, for steering the model's hidden states away from a hallucinatory state. 
However, common studies employ different setups and do not properly separate different possible causes of hallucinations, making interventions misguided.  
In this work, we introduce a method for categorizing examples based on the model's prior knowledge, named \bn. We construct \bn benchmarks that support interventions in two settings: open-book and closed-book question answering. 
Using the benchmarks, we perform an extensive investigation of the effect of different choices for intervention, such as the intervened components, and how often and how strongly to intervene. We find that intervention success varies depending on the component, with the attention blocks performing well and the residual stream proving detrimental to language modeling capabilities. We also show that interventions can benefit from representative vectors collected before, rather than after, a hallucination occurs. Finally, we introduce a new dynamic intervention, which intervenes only if needed, and thus is more robust than standard static interventions.\footnote{Code and datasets at \url{https://github.com/technion-cs-nlp/hallucination-mitigation}.}

\end{abstract}

\section{Introduction}

A notable issue identified in LLMs is their prone to generate outputs that are not grounded in the input of the model or in real-world facts, or outputs that are inconsistent with earlier generations in a session \citep{survey_of_hallucination_in_natural_language_generation, Towards_understanding_sycophancy_in_language_models, Calibrated_language_models_must_hallucinate}. These different issues are typically titled collectively as \emph{hallucinations}, which are important to address as they reduce the reliability of LLMs. Numerous efforts have been made to tackle these issues, including blackbox methods, which fix the generated text by supervised and reinforcement learning finetuning recipes \citep{Fine-tuning_Language_Models_for_Factuality,Info_for_Faithfulness,rlhf}, alongside whitebox methods, which aim to detect and prevent hallucinations using directly the model's inner states \citep{CCS, LLM_Polygraph, Weakly-Supervised_Detection_of_Hallucinations_in_LLM_Activations,iti, geometry_of_truth, Representation_engineering_lorra}. 

In this work, we focus on the whitebox approach, which (a) may enable earlier actions upon hallucination during generation; and (b) provides a better understanding of the hallucination mechanism in LLMs.
We first categorize the following LLM knowledge types given a question: (1) the model does not know the correct answer (2) the model considers a few options among the correct answer (3) the model knows the correct answer.
We claim that a proper benchmark for this task should only include cases of type 3, as hallucinations in this knowledge type could potentially be mitigated without external knowledge.
We then propose a method to automatically differentiate between these types for a given model, constructing model-specific benchmarks, where a model has a \textbf{W}rong \textbf{A}nswer despite having \textbf{C}orrect \textbf{K}nowledge (WACK).
The automation is done by repeat sampling from the model. The final dataset uses only type 3 and labels automatically hallucination/grounded based on the model's preference.

We present \bn hallucination benchmarks in two settings, one that requires the model to rely on knowledge from the input context in the prompt and not its parametric knowledge (\emph{open-book}) and one that requires the model to rely only on its parametric knowledge (\emph{closed-book}). 
The closed-book setting is intriguing, wherein the model hallucinates despite knowing the correct answer and lacking any alternative answer in the context. Intuitively, one might assume that this should not occur. However, recent work shows how user input can trigger such behavior \citep{The_Waluigi_Effect,theoretical_Waluigi_Effect}. We propose a novel setting to examine this by adding irrelevant text with minor mistakes at the start of the prompt, thus triggering a behavior similar to the one a user may create.

Armed with tailored benchmarks, we turn to analysis and utilize steering vectors \citep{steering_vectors,steering_vectors2} to mitigate hallucinations, by adding a vector to a hidden state to steer away from hallucinations. We target detection before hallucination onset, intending to steer the generation away from hallucination before it is materialized in surface tokens. Unlike prior work, which detected and created a steering vector for hallucinations after they were generated \citep{iti,geometry_of_truth,Representation_engineering_lorra,trfr,nl-iti}.
Despite some success in mitigation in prior work, there are no clear guidelines on how to intervene. We use our framework and benchmarks to create such guidelines by rigorously
investigating different choices for intervention such as which component in the model’s architecture to apply it (MLPs, attention blocks, residual stream, and specific heads). We introduce dynamic intervention, intervening only when needed, and compare it to standard static intervention. 
We discover several insights: (i) intervening using pre-hallucination inner states for the steering vector is better than post-hallucination; (ii) intervention on the residual has a model's performance compromise, which is mitigated to some extent when intervening dynamically; (iii) intervening on the attention component is preferred over the MLP part; (iv) model preference between answers (hallucination vs. grounded), which is commonly used as an evaluation metric, is not a good approximation for whether the model will generate the grounded answer;

Our main contributions are: 
(\RomanNumeralCaps{1})  A methodology (\bn) for constructing per-LLM benchmarks containing grounded and hallucination outputs in cases where the model knows the information for a correct response in its parameters, in both open-book and closed-book settings. We released the datasets for our tested models (Llama2-7B and Goat-7B);
(\RomanNumeralCaps{2}) A comprehensive investigation into optimal intervention strategies within the model architecture;
(\RomanNumeralCaps{3}) A novel \textit{dynamic} intervention approach that intervenes only when needed, thus providing a robust solution.

\section{Problem setup}\label{sec:problem_setup}

Consider a question-answering (QA) setup, where a model is given a question and generates an answer.
We consider two settings: open-book QA where the gold-truth answer appears in a given context and closed-book QA where the gold-truth answer resides in the model's parameters.
A common recipe for mitigating hallucinations involves the following steps: (1) constructing a dataset with hallucination labels; (2) training a detector (classifier) on intermediate representations to identify cases of hallucinations; and (3) mitigation - intervening in the model's computation to prevent a hallucinate answer. The third step requires the detector from step two to decide where to intervene. In this work we focus on steps (1) and (3). See Appendix \ref{appendix:Detection results} for details on how we perform step (2), as well as various detection results.

\subsection{Mitigation setup}
\label{ssec:mitigation_setup}

We aim to mitigate hallucinations by adjusting the model's activations. Following prior work in this area \citep{iti,nl-iti,trfr,geometry_of_truth,Representation_engineering_lorra} and in other tasks \citep{bau2018identifying,ravfogel-etal-2021-counterfactual}, our intervention approach is to add a vector $d_{l, c}$ to the model's activations $v_{l, c}$, at some layer $l$ and component $c$, where   $c \in \{\text{MLP},\text{Attention},\text{Head},\text{Residual}\}$.  For the MLP, attention block, and residual, our modification is at the output of the layer, and for the heads, it is done before applying the projection matrix back to the embedding space.
The vector $d_{l, c}$ defines a linear separation between hallucination and grounded subspaces. Then, during inference, our intervention formula, $v_{l, c} = v_{l, c} + \alpha*d_{l, c}$, shifts the activation vector in the direction of the grounded subspace, aiming at making the model generate a grounded answer. Here $\alpha$ emphasizes the impact of the intervention. 
$d_{l,c}$ is sometimes known as a steering vector  \citep{steering_vectors2,steering_vectors}.
Our work explores the gory details of the mitigation step, how to compute the steering vector, evaluate its performance, and suggest insights for efficient mitigation.
Notably, steering vectors do not introduce supplementary knowledge to the model, and computing them typically relies on labeled data. We use this to decide how to construct the dataset.

\subsection{Knowledge categorization}
\label{ssec:eval_setup}

In closed-book setting, Hallucinations can be caused by different reasons, depending on whether or not the model \emph{knows} the golden response, i.e., the model is expected to induce the golden response tokens from its activations with high confidence. 
We define three types of knowledge categories that in each hallucination can occur:
\textbf{(1)} The model does not know the golden answer at all.
\textbf{(2)} The model considers the golden answer as one of a few possible ones, without a clear preference for the golden answer. 
\textbf{(3)} The model knows and has a clear preference for the golden answer.

We argue that if the model does not know the golden answer (type 1), it is possible to prevent hallucination by using external knowledge or additional training. If the model has no clear preference for the golden answer (type 2), intervention without external knowledge might be possible but seems unlikely. Instead, the model could refrain from answering in cases of uncertainty in the answer. This is a different type of intervention that we defer for future work.
On the other hand, if the model does know the golden answer, it may be possible to direct its output toward the golden answer via whitebox intervention, without additional knowledge.
Therefore, we focus on \emph{type 3} in this work. The automatic way we differentiate between those types of knowledge is explained in Section \ref{sec:dataset_creation}.

This is a novel categorization of hallucination based on knowledge that was not done in prior work on mitigation \citep{truthx,iti,geometry_of_truth,Representation_engineering_lorra,trfr,nl-iti}, and therefore prior benchmarks mixed all types.\footnote{An exception is a concurrent work \citep{whispers}, which accounted for the model's knowledge in dataset creation.} We advocate for the significance of this categorization to address the diverse causes of hallucination effectively.

\begin{figure*}
\centering
  \centering
  \includegraphics[width=0.59\linewidth]{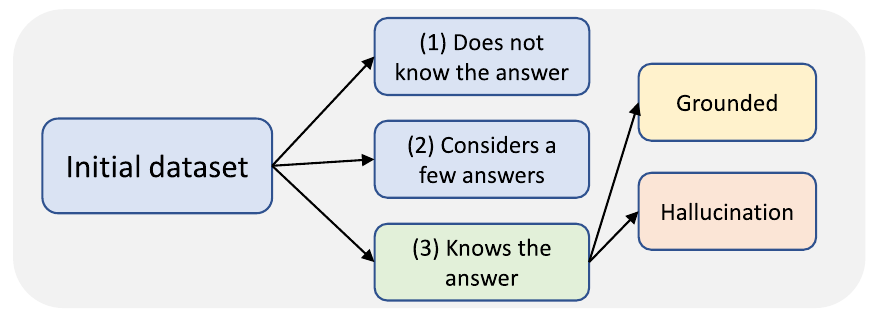}
\caption{Outline of \bn method for dataset construction.}
\label{fig:Outline of our dataset creation.}
\end{figure*}
\section{\bn method for dataset construction}\label{sec:dataset_creation}

Training and evaluating detectors and steering vectors for mitigating hallucinations necessitates labeled datasets, containing hallucination and grounded examples.
We aim to construct such datasets for type 3 knowledge (Section \ref{ssec:eval_setup}). To this end, we describe an automatic method to differentiate between the three knowledge types. Then, we create a labeled hallucination dataset in open-book and closed-book QA settings. Figure \ref{fig:Outline of our dataset creation.} outlines this process.

\subsection{Knowledge example selection}\label{sec:Hallucination Example Selection}

To automatically differentiate between the three types of knowledge situations that can create hallucinations, we use repeated sampling from a tested LLM $M$. Concretely, given a QA example consisting of a question and a gold answer, we sample $K = 5$ answers from $M$ using temperature sampling, and categorize the example according to the number of times the gold response was sampled: 
0 times refers to type-1, 1--3 times refers to type-2, and 4--5 times refers to type-3. 
We retain only type-3 examples, thus tailoring a dataset per model $M$. henceforth in the paper, we will use the terminology for type-3 examples as \emph{the model knows the answer}.
From those examples, we will next create a labeled hallucination dataset. This method is easy to implement and we found it useful for splitting into our three types.\footnote{One may also train a dedicated model to separate examples by model's knowledge  \citep{p(ik)}, but that is more complicated than our approach.} We applied it in both open-book and closed-book settings; see below. This is primarily needed in the closed-book setting but for consistency, we applied it in both settings. See Appendix \ref{appendix:General Dataset Construction Specifics} for specifics.

\subsection{Open-book dataset creation} \label{sec:data-context}

In this scenario, we study hallucinations that occur when there is a conflict between a model's parametric knowledge and a given context. A hallucination in this case is when the model's answer deviates from information given in the \emph{prompt context}. 
We use the DisentQA dataset \cite{disentqa}, where each example is a question \question, a text context \context, and two answers: a parametric answer \parametric and a contextual answer \contextual. See Figure \ref{fig:disentqa-example} for an example. 

\begin{figure}
    \centering
     \fbox{\footnotesize\parbox{\textwidth}{
    \textbf{question}: how many chromosomes does a human diploid cell have?\\context:  Humans are diploid organisms , carrying two complete sets of chromosomes : one set of 23 chromosomes from their father and one set of 23 chromosomes from their mother . The two sets combined provide a full complement of 2 chromosomes . This total number of chromosomes is called the chromosome number . The zygotic number is defined as the number of chromosomes in zygotic cells . Human zygotes are diploid , hence with a zygotic number of 2 . \\\textbf{answer}:
    \\\textbf{Parametric answer} (\parametric): 46. \textbf{Contextual answer} (\contextual): 2
    }}
\caption{Example from DisentQA (open-book) with both contextual and parametric answers.}
\label{fig:disentqa-example}
\end{figure}

We keep only examples for which the model \textit{knows} the parametric answer, meaning it was generated at least 4 out of 5 times when sampling given only the question \question without the additional context. 

Next, given a full DisentQA example \{$q, c, a_p, a_c$\}, containing a question $q$ and context $c$ inputs, and the parametric $a_p$ and contextual $a_c$ gold answers, we measure how much the model prefers $a_c$ over $a_p$: \score $= \text{Rank}(a_p \mid q, c) - \text{Rank}(a_c \mid q, c)$. Where $\text{Rank}(a)$ represents the rank assigned by the model to token $a$ compared to all other tokens after projection to the vocabulary space at the last layer; a low-rank number indicates a higher preference for answer $a$.
If  \score $\geq T$ for some threshold $T$ then the example is labeled `grounded', and if \score $\leq -T$ it is labeled `hallucination'. 
The result of this process is a model-specific open-book dataset labeled for hallucinations. See details in Appendices \ref{appendix:General Dataset Construction Specifics}, \ref{appendix:Open-Book Dataset Construction Specifics}, and \ref{appendix:Datasets Statistic}.

\subsection{Closed-book dataset creation}\label{subsec:Closed-Book Dataset Creation}

In this setting, we study hallucinations that occur when an LLM generates an answer that deviates from its parametric world knowledge.
To this end, we follow prior work that shows that models may simulate different personas, making them hallucinate even though they know the answer \citep{theoretical_Waluigi_Effect,Personas,simulators}. 
This kind of behavior may happen when a user writes something untrue or misleading \citep{The_Waluigi_Effect, theoretical_Waluigi_Effect, hallucination_snowball,flat_earth}.

Concretely, we constructed 20 \emph{bad-shot examples} using ChatGPT \citep{rlhf}. These are false QA pairs, where the false answer is semantically similar to the gold one. Thus, simulating minor mistakes that a user may create.\footnote{\cite{yao2023llm} generated hallucination trigger prompts by constructing them to be out-of-distribution or semantically weak. In contrast, we believe that our bad-shots prompts maintain greater coherence and alignment within the distribution.} 
For instance, here is a \emph{good-shot} example and its corresponding \textit{bad-shot} example: 
\begin{description}[itemsep=0pt,topsep=1pt,parsep=1pt]

    \item[Good-shot:] \underline{question}: Which element has the chemical symbol 'H'? \underline{answer}: Hydrogen
    \item[Bad-shot:] \underline{question}: Which element has the chemical symbol 'H'? \underline{wrong answer}: Helium
\end{description}

\begin{figure}
    \centering
     \fbox{\footnotesize\parbox{.75\textwidth}{\textbf{few shots:}
    3 good-shots (or 3 bad-shots)\\
    \textbf{question:} Which colour lies between red and yellow in the visible spectrum? \\\textbf{(wrong) answer}:
    \\\textbf{Golden answer} (\golden): Orange
    }}

\caption{Example from TriviaQA (closed-book), with the addition of good-shot or bad-shot at the beginning of the prompt and using \emph{wrong answer} instead of \emph{answer} in the bad-shot permutation. }

\label{fig:triviaqa-example}
\end{figure}

Then, for a given LLM $M$, we used $M$ to answer the questions in TriviaQA \citep{triviaqa}  in a closed-book setup. As part of the question instruction, we provide the 3 good-shots or 3 bad-shots as few-shot exemplars for the task (Figure \ref{fig:triviaqa-example}).
For a given example \{$q$, $a_g$\} of a question $q$ and a gold answer $a_g$, we label it as `grounded' or `hallucination' via the difference in rank of $a_g$ following the bad-shot or good-shot prompts: \score $= \text{Rank}(a_g \mid \text{bad-shots}) - \text{Rank}($\golden $\mid \text{good-shots})$.
If the rank does not change (\score $= 0$), the example is considered grounded.
If the $\text{Rank}(a_g \mid \text{bad-shots})$ worsens (\score $\geq 1$) and the model predicted the answer correctly given the good-shot, we consider the example as a hallucination.
Otherwise, the example is removed. 
This labeling procedure is illustrated in Figure \ref{fig:bad_shot_hall}. 
For more details and examples of hallucinations with bad shots, see  Appendices \ref{appendix:General Dataset Construction Specifics}, \ref{appendix:Closed-Book Dataset Construction Specifics} and \ref{appendix:Datasets Statistic}.

\begin{figure*}
\centering
  \centering
  \includegraphics[width=\linewidth]{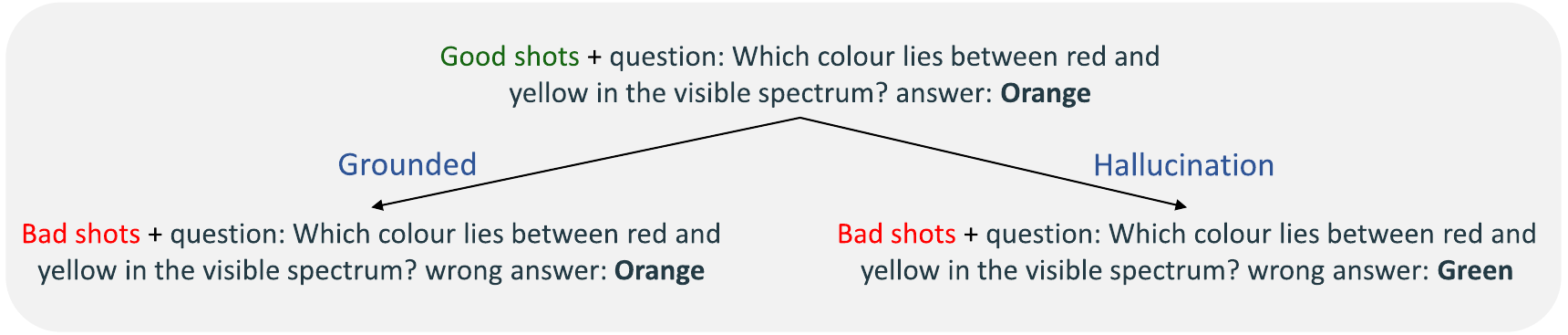}

\caption{Hallucination labeling in the closed-book setting. Model generations are in \textbf{bold}.}

\label{fig:bad_shot_hall}
\end{figure*}

\section{Intervention analysis approach}
\label{sec:approach}

With our \bn datasets, using knowledge of type 3, where hallucinations occur although the model knows, we perform a rigorous analysis of whitebox model intervention via steering vectors (Section \ref{ssec:mitigation_setup}). 
We discuss the design choices of intervention and reassess proper evaluation metrics.

\begin{figure*}
\centering
  \centering
  \includegraphics[width=\linewidth]{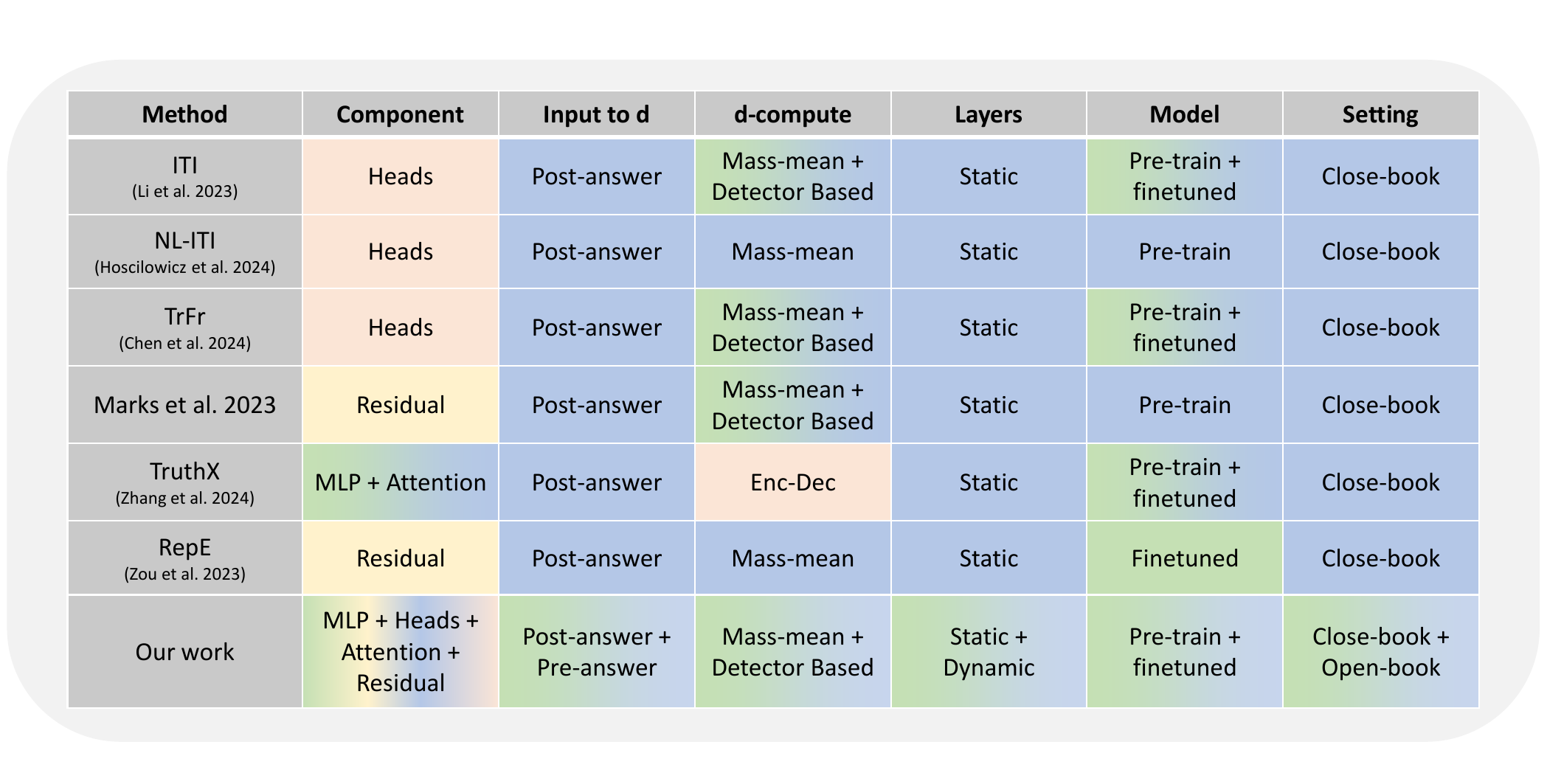}

\caption{Methodological variability in Intervention with a list of the related papers and what variables each uses. Dynamic is a new methodology we show in this work.}

\label{fig:Exploring Methodological Variability in Intervention.}
\end{figure*}
\subsection{Exploring methodological variability in intervention}\label{sec:intervention_method}

Various studies employed different settings for intervention, and it remains unclear which approaches are most effective. We conduct experiments with the common choices that appear in most works to establish guidelines and introduce novel settings. Figure \ref{fig:Exploring Methodological Variability in Intervention.} presents the variables we explore.
Specifically, we identify the following degrees of freedom from prior work in designing an intervention. 

First, %
a steering vector may intervene in different \textbf{components}: MLPs \citep{truthx}, attention blocks \citep{truthx}, heads \citep{trfr,iti,nl-iti}, or residual stream \citep{geometry_of_truth,Representation_engineering_lorra}.\footnote{\cite{Representation_engineering_lorra} also used the attention for a different intervention using LoRA \citep{hu2021lora}.} In this work, we explore all of the components.

Second, the steering vector $d_{l, c}$ is commonly computed using examples containing the generated answer \citep{iti,nl-iti,trfr,geometry_of_truth,truthx,LLM_Polygraph,Representation_engineering_lorra}. However, it is then used for mitigation \emph{before} the answer is generated. Thus, there may be a discrepancy between the way it was computed and how it is used. 
In our experiments, we consider both the common \textbf{post-answer} method and a new \textbf{pre-answer} setup where the steering vector is calculated \emph{before} answer generation instead of after. 
Pre-answer is a novel setting we introduce, which is more aligned with how $d$ is used during intervention.

Third, the intervention (\textbf{$\mathbf{d}$-compute}) may be computed from a detection classifier or another summary statistic of hallucination vs.\ grounded cases. In particular, it is common to take the linear separator for the former \citep{geometry_of_truth,trfr,iti} and the difference in mean vectors of grounded and hallucination examples for the latter \citep{iti,geometry_of_truth,Representation_engineering_lorra,trfr,nl-iti}.\footnote{One exception is \citet{truthx}, who trained an auto-encoder to learn $d$.} In both cases, the detector makes an intervention decision.

Additionally, we introduce another degree of freedom. \textbf{Static} intervention occurs in the same components and layers, for all examples, usually by picking the places that got the highest detection score. This is a common strategy \citep{iti,geometry_of_truth,truthx,LLM_Polygraph,trfr,nl-iti}. However, such drastic intervention may not always be necessary. 
We suggest a different option: \textbf{dynamic} intervention, in which the components to intervene are chosen in each example differently. Our intuition is that intervention should be applied when hallucination is likely to occur, which is pointed at by our detector. 
Thus, given a threshold $TD$, we intervene only in components whose detector accuracy is above $TD$ (a high-quality detector) and the detector classified the example as a hallucination for them.

We also explored the difference in intervention effect between fine-tuned and pre-trained models and investigated both open-book and closed-book settings.

\subsection{Evaluating intervention impact}\label{sec:evaluation}
Recent work lacks a consistent evaluation method for measuring intervention success. Most compare probabilities of the golden answer and other possible answers \citep{geometry_of_truth,Representation_engineering_lorra,iti,truthx,trfr,nl-iti}, while others also evaluate generation with a language model \citep{iti,truthx,trfr}. 
We advocate for examining multiple metrics to provide a holistic picture of intervention effects. 

To measure intervention success, we report the \textbf{classification accuracy}. In the open-book case, we measure how often the model prefers the contextual answer over the parametric answer by comparing Rank(\contextual)$\le$Rank(\parametric).
In the closed-book case, we compare the rank of the golden answer given a bad-shot prompt and a good-shot prompt (Rank$($\golden $\mid$ bad shots$)$$\le$Rank$($\golden$ \mid$ good shots$)$).

The classification accuracy is simple and intuitive. However, recent work suggests that comparing probabilities of answer tokens is not a good proxy for text generation \citep{generation_vs_classification}. 
We argue that this discrepancy is particularly crucial for intervention assessment.
Therefore, we report the \textbf{generation accuracy}, which examines whether the desired answer was generated in greedy decoding of 20 tokens. Note, that we focus on a generation of a few tokens, as the model may produce a lengthy correct answer that should also be classified as grounded.

Lastly, to ensure that the intervention has not compromised the model's performance, 
we report \textbf{Wiki ppl}, the perplexity on 100 random articles from WikiText-103 \citep{wikitext_paper}. 
See details in Appendix \ref{Specific_Evaluation_Measures+On_no_context_setting}. \footnote{We also tried pure perplexity but it was less stable than Wiki ppl.}

\section{Implementation details}\label{sec:Implementation Details}

In all experiments, we randomly selected 1000 examples from each label (\textit{hallucination}/\textit{grounded}) for analysis in each dataset %
and split to 70\%/10\%/20\% for training/validation/test.
We use a linear SVM classifier for detection, as in prior work  \citep{iti,geometry_of_truth}.
Each experiment was repeated with three random seeds for the SVM and split into training/validation/test sets. We report average results with standard deviations.
Datasets construction and all the results ran on NVIDIA RTX 6000 Ada (49GB) with 4 CPUs. 
Each dataset construction and each result takes approximately 12 hours (30 GPU days).

In addition to evaluating Llama-2-7B \citep{llama2} on all experiments, we evaluated Goat-7B \citep{goat_model} on a subset of the experiments 
to highlight performance differences between base and fine-tuned models.\footnote{We leave chat models for future work as they modify the prompt to a chat format that makes direct comparisons of base and chat models difficult.} 

To explore the various options discussed in Section \ref{sec:intervention_method}, we start with dynamic pre-answer mass-mean direction intervention with $\alpha=5$, a configuration that showed promising results in initial exploration. We then modify a single intervention variable at a time (dynamic/static, pre-answer/post-answer, etc.) to assess its impact on performance in an ablation setup.
Additional hyperparameter details, ablation choices, and results on the general distribution of the data are in Appendices \ref{appendix:hyperparams}, \ref{appendix:classifier_vs_mass_mean} and \ref{appendix:wighted_results_all_data}.

\section{Intervention results}\label{sec:Recipe_for_Intervention}

\begin{figure}
\centering
 \centering
\begin{subfigure}[b]{0.33\textwidth}
  \centering
  \includegraphics[width=\linewidth]{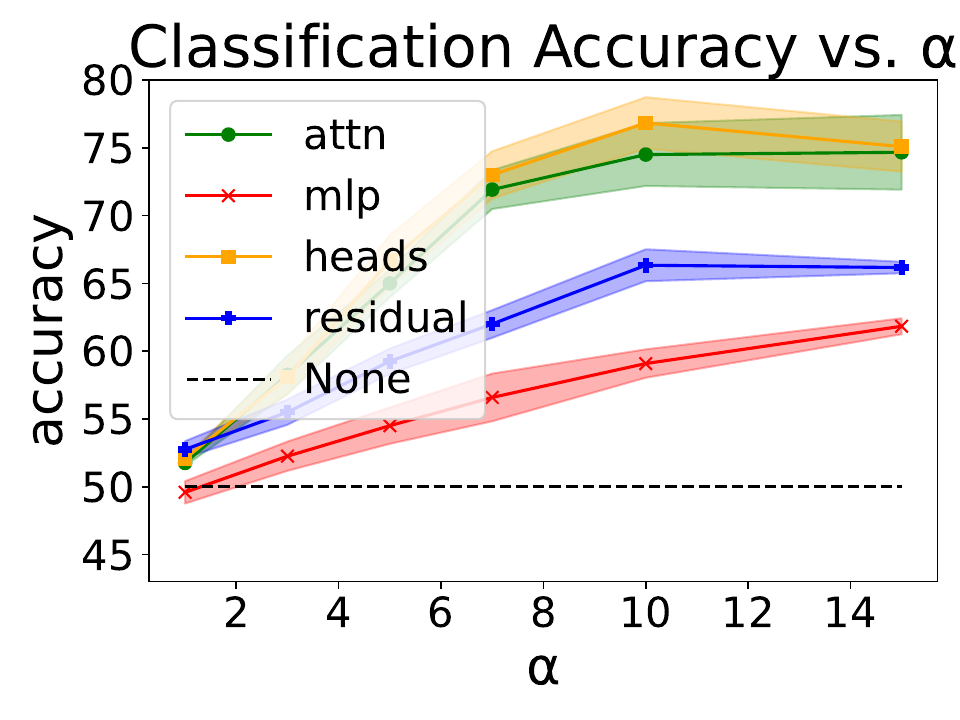}
  \caption{}
  \label{fig:classification_acc_on_different_alphas}
 \end{subfigure}%
 \hfill
 \begin{subfigure}[b]{0.33\textwidth}
  \centering
  \includegraphics[width=\linewidth]{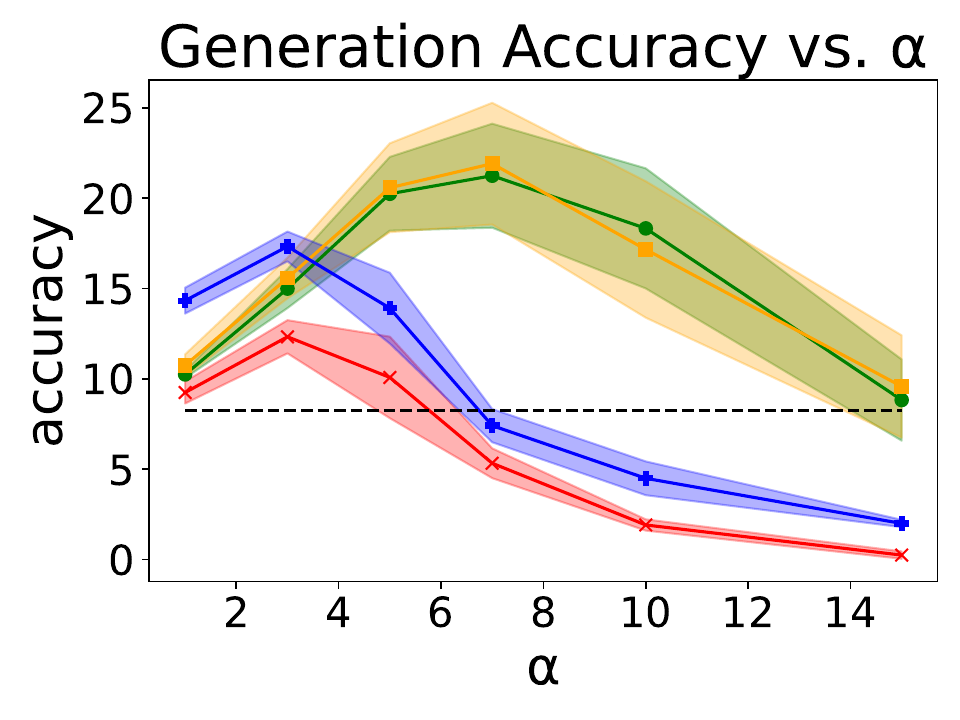}
  \caption{}
  \label{fig:generation_acc_on_different_alphas}
 \end{subfigure}
 \hfill
 \begin{subfigure}[b]{0.33\textwidth}
     \includegraphics[width=\textwidth]{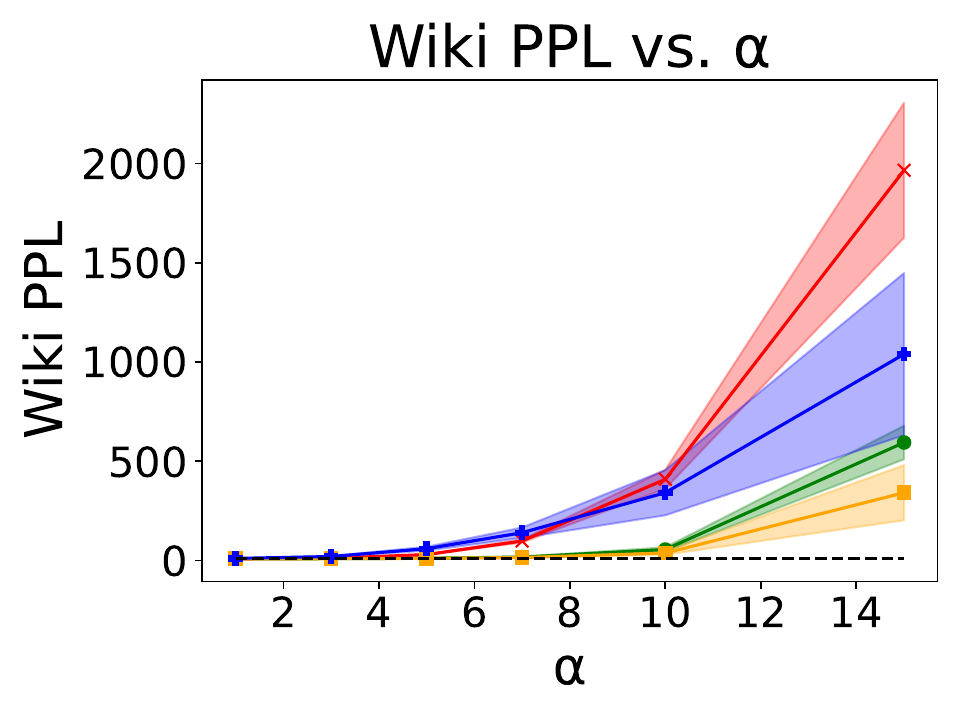}
     \caption{}
     \label{fig:wikipedia_pp_on_different_alphas}
 \end{subfigure}

 \caption{Evaluation methods across all components and alphas on open-book DisentQA-\bn.}

 \label{fig:Evaluation methods across all components and alphas}
\end{figure}

\begin{figure}
\centering
\begin{subfigure}[b]{0.33\textwidth}
  \centering
  \includegraphics[width=\linewidth]{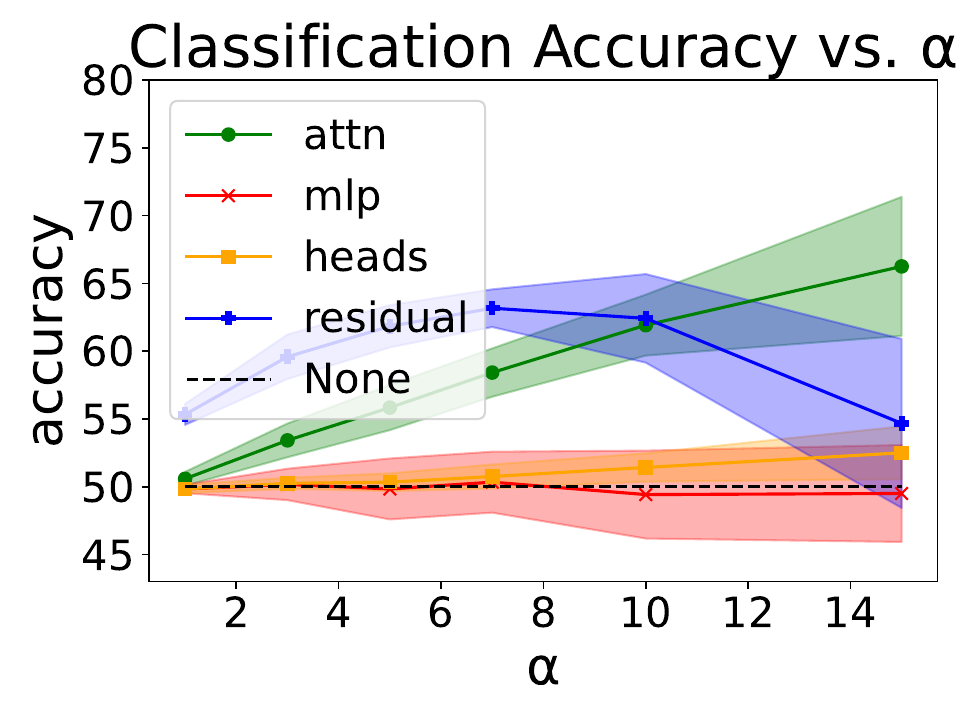}
  \caption{}
  \label{fig:classification_acc_on_different_alphas_trivia}
 \end{subfigure}%
 \hfill
 \begin{subfigure}[b]{0.33\textwidth}
  \centering
  \includegraphics[width=\linewidth]{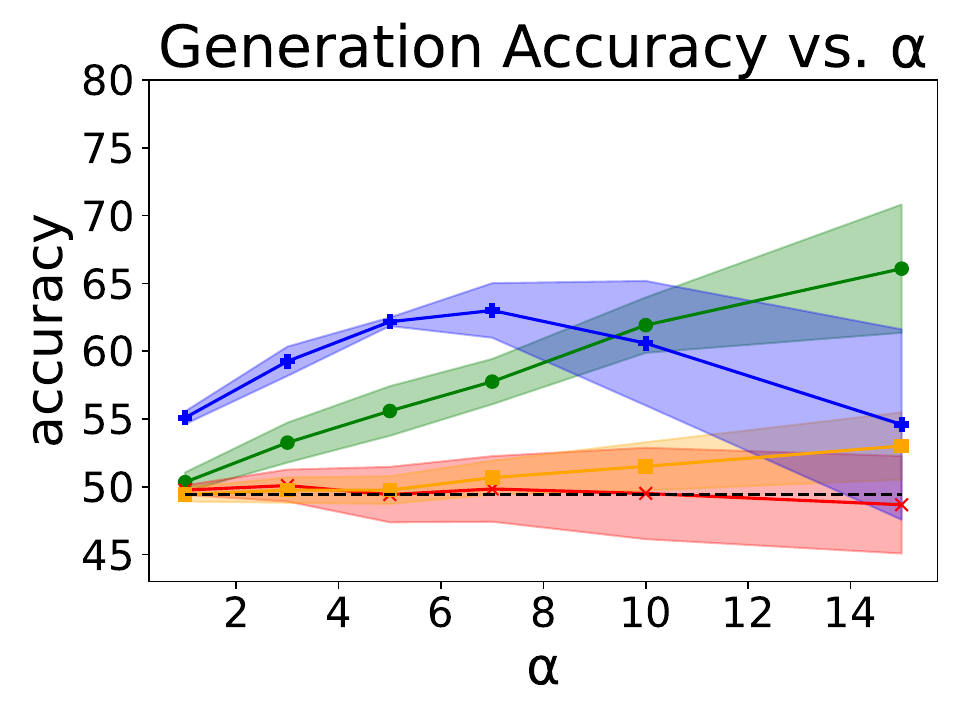}
  \caption{}
  \label{fig:generation_acc_on_different_alphas_trivia}
 \end{subfigure}
 \hfill
 \begin{subfigure}[b]{0.33\textwidth}
     \includegraphics[width=\textwidth]{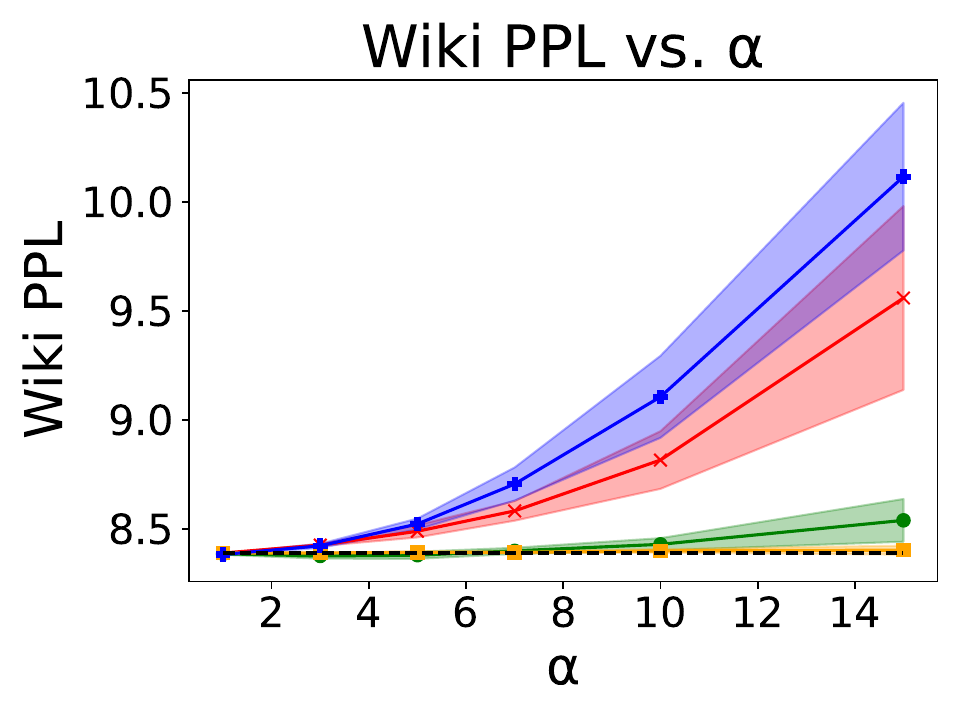}
     \caption{}
     \label{fig:wikipedia_pp_on_different_alphas_trivia}
 \end{subfigure}

 \caption{Evaluation methods across all components and alphas on closed-book TriviaQA-\bn.}
 \label{fig:Evaluation methods across all components and alphas using Llama2 model on TriviaQA-no-context dataset}
\end{figure}

Figures \ref{fig:classification_acc_on_different_alphas}, \ref{fig:generation_acc_on_different_alphas}, \ref{fig:classification_acc_on_different_alphas_trivia} and \ref{fig:generation_acc_on_different_alphas_trivia} show the accuracy on the open-book DisentQA-\bn-Llama2 (DisentQA-\bn for short) and closed-book TriviaQA-\bn-Llama2 (TriviaQA-\bn for short) settings w.r.t intervening with increasing values of $\alpha$, which controls the magnitude of the intervention.
While in \cbQA setting, classification and generation accuracy are coupled, in \obQA, when $\alpha$ is increased, classification accuracy mostly increases but generation peaks and then drops. This indicates that classification accuracy is not always a good approximation for generation accuracy, which is what a user actually receives from an LLM. We also see an increase in Wiki ppl as $\alpha$ grows (Figures \ref{fig:wikipedia_pp_on_different_alphas} and \ref{fig:wikipedia_pp_on_different_alphas_trivia}), pointing at worse language modeling ability with stronger intervention. This occurs in both settings but with different impacts per intervened component. Thus, evaluating also perplexity is recommended to assess intervention performance.

Component-wise, in \obQA, \textit{attention} and \textit{heads} are best performing and very similar, under classification, generation, and perplexity evaluation. On the other hand, in \cbQA, \emph{residual} has better classification and generation accuracy, but declining at higher $\alpha$, and perplexity levels are increasing quickly too, while \emph{attention} accuracy keeps improving without any perplexity drop.

\begin{figure*}[t]
\centering
\begin{subfigure}{0.33\textwidth}
  \centering
  \includegraphics[width=.95\linewidth]{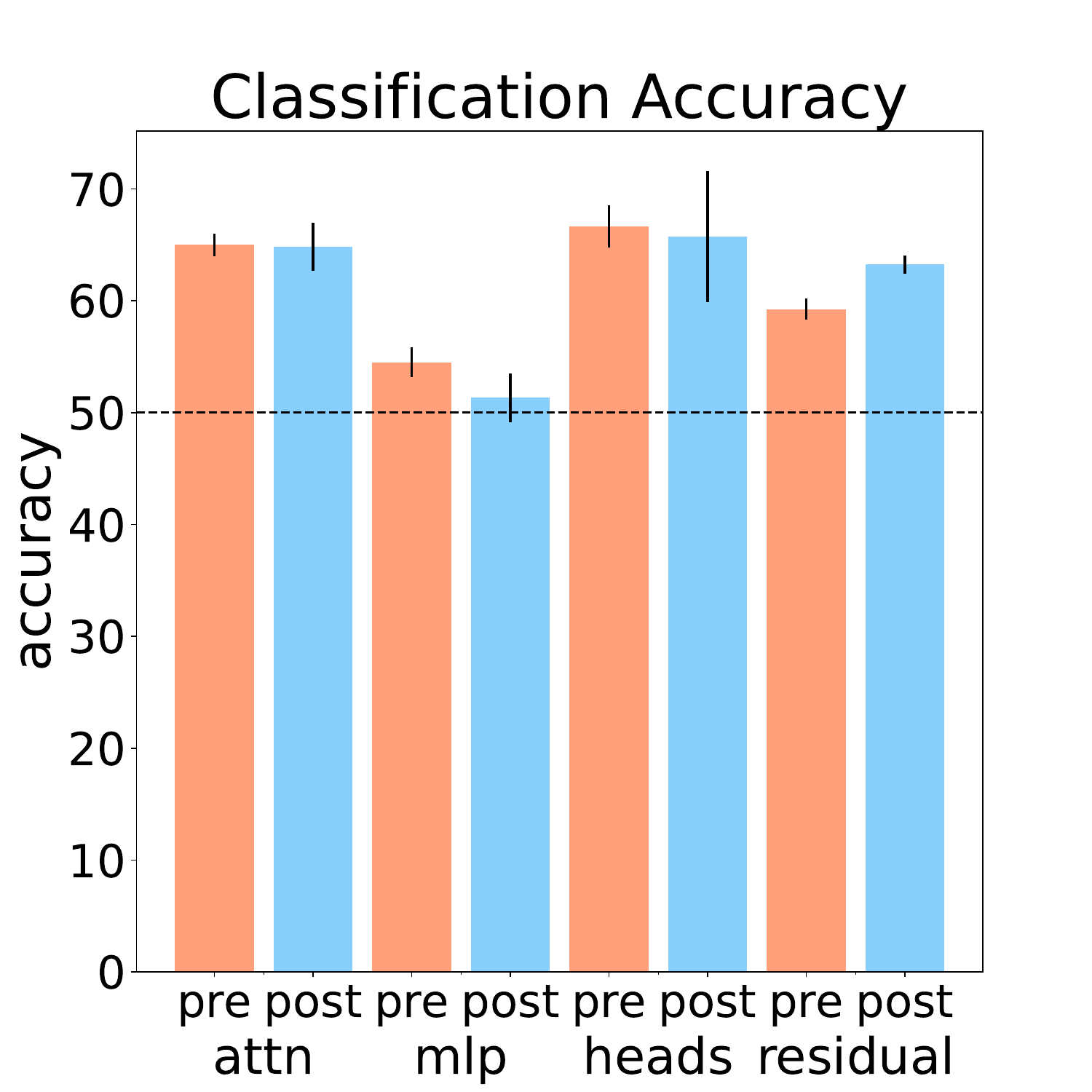}
  \caption{}
  \label{fig:pre_post_classification}
 \end{subfigure}%
 \hfill
 \begin{subfigure}{0.33\textwidth}
  \centering
  \includegraphics[width=.95\linewidth]{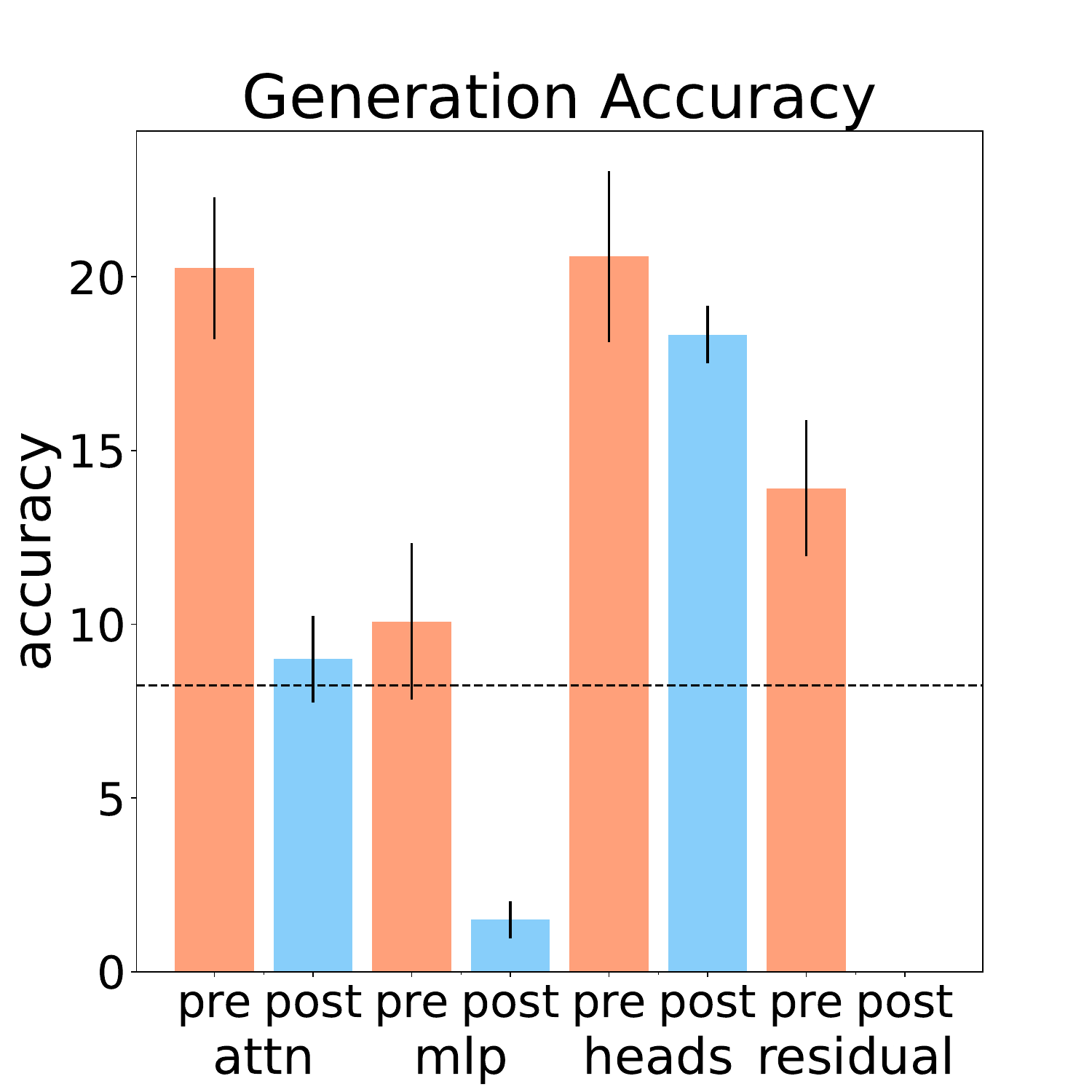}
  \caption{}
  \label{fig:pre_post_generation}
 \end{subfigure}
  \hfill
 \begin{subfigure}{0.33\textwidth}
  \centering
  \includegraphics[width=.95\linewidth]{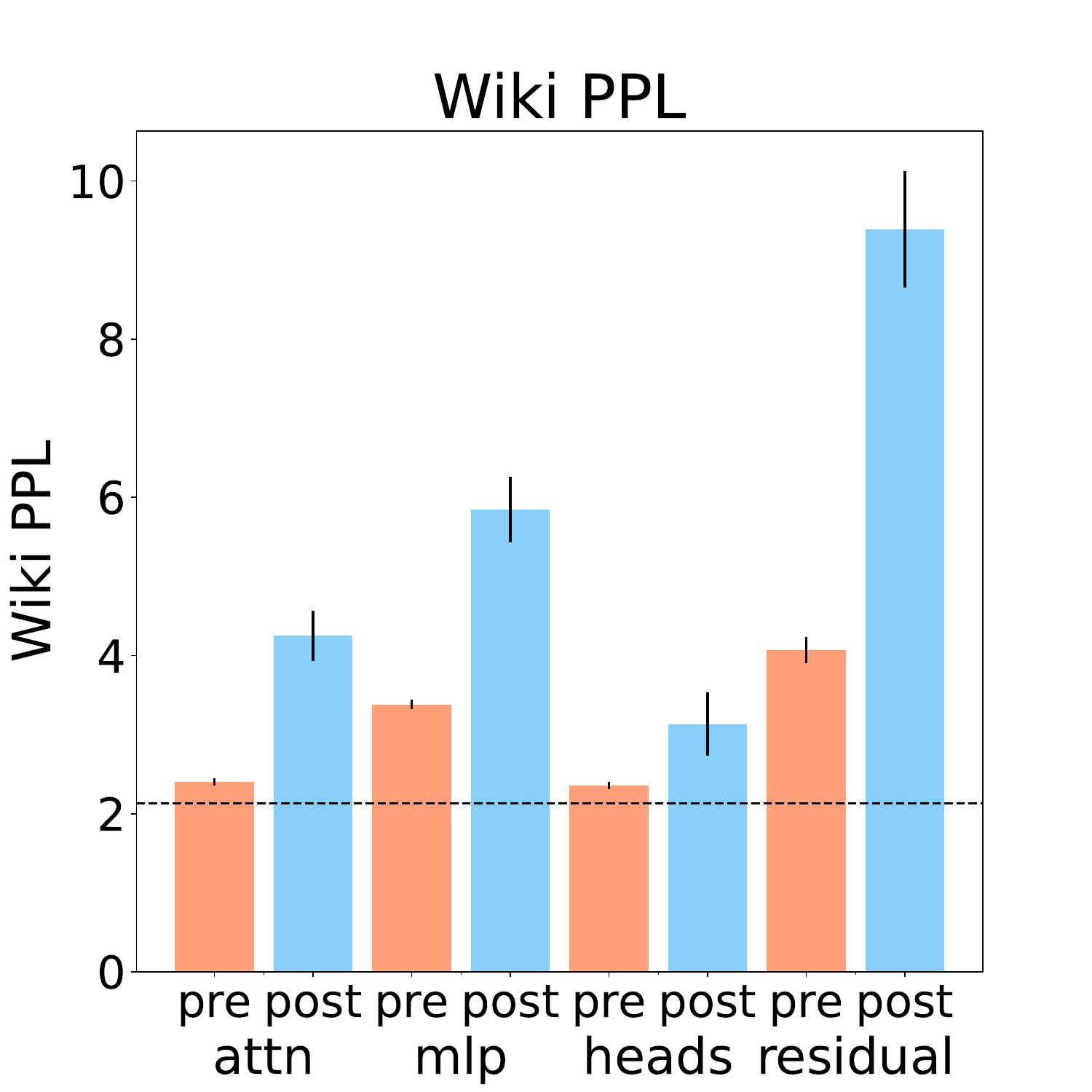}
  \caption{}
  \label{fig:pre_post_wiki_pp}
 \end{subfigure}

\caption{Pre- vs. post- answer on DisentQA-\bn (open-book) results.
PPL is on a log scale.}

\label{fig_pre_post}
\end{figure*}

In Section \ref{sec:intervention_method} we proposed a novel \emph{pre-answer} intervention strategy. Figure \ref{fig_pre_post} presents a comparison between pre-answer and post-answer interventions.\footnote{We run it only on the open-book setting as it has a clear answer (\contextual for grounded and \parametric for hallucination).}
We can see that while the two strategies are on par on classification accuracy, regarding generation, and perplexity, it is clear that \emph{pre-answer} is a better strategy. In addition, under \emph{post-answer}, intervention at the \emph{heads} component is best performing, but under \emph{pre-answer}, \emph{attention} component gains and matches \emph{heads} performance.

We conclude that a recommended intervening component in both settings is the \emph{attention}, and to reach good results intervention should be performed using the \emph{pre-answer} strategy.

\subsection{Static interventions vs. dynamic interventions}\label{sec:static_vs_dynamic}

Static intervention, which modifies the same components for every input and typically targets areas with the highest detection, is the common approach in the literature.
Dynamic intervention, which we propose, decides which components to modify given an input example based on the detector's decision. Dynamic intervention might be preferable since unnecessary interventions might do more harm than good, e.g., hurting generation abilities without improvement to ``correctness''.

Table \ref{Dynamic intervention vs. static intervention on DisentQA} shows a minimal difference between the two approaches in most cases with a slight overall preference for the static intervention. However, when intervening in the residual, dynamic intervention yields significantly better results, while static intervention leads to poor generation accuracy and perplexity.
Additionally, examples of generated text (in Appendix \ref{appendix:generation_static_synamic}) illustrate how static intervention in the residual component adversely affects generation. We conclude that when intervening in the residual, it is preferred to use the dynamic intervention.

\begin{table*}[t!]
  \caption{Dynamic vs.\ static intervention.}

\centering
\small
  
  \begin{tabular} {p{0.08\linewidth}p{0.07\linewidth}p{0.1\linewidth}p{0.1\linewidth}p{0.1\linewidth}p{0.1\linewidth}p{0.1\linewidth}p{0.1\linewidth}}

\toprule 
\multicolumn{2}{c}{Setting} & \multicolumn{3}{c}{DisentQA-\bn} & \multicolumn{3}{c}{TriviaQA-\bn}\\
\cmidrule(lr){1-2} \cmidrule(lr){3-5} \cmidrule(lr){6-8}

& & Classification & Generation & & Classification & Generation &  \\ 
& & \multicolumn{1}{c}{Accuracy} & \multicolumn{1}{c}{Accuracy}& Wiki ppl &  \multicolumn{1}{c}{Accuracy} &  \multicolumn{1}{c}{Accuracy} & Wiki ppl \\
\midrule 

\multirow{2}{*}{Attention}&static& $\mathbf{65.33_{\pm4.4}}$ & $\mathbf{23.0_{\pm1.97}}$ & $\mathbf{8.97_{\pm0.12}}$ & $\mathbf{56.25_{\pm3.69}}$ & $\mathbf{57.0_{\pm3.68}}$ & $\mathbf{8.33_{\pm0.08}}$ \\
&dynamic & $65.0_{\pm1.02}$ & $20.25_{\pm2.04}$ & $11.04_{\pm0.51}$& $55.83_{\pm1.66}$ & $55.58_{\pm1.83}$ & $8.38_{\pm0.02}$ \\

\midrule

\multirow{2}{*}{MLP}&static& $51.08_{\pm1.36}$ & $\mathbf{12.42_{\pm1.94}}$ & $\mathbf{9.78_{\pm0.56}}$  & $49.58_{\pm2.97}$ & $\mathbf{50.08_{\pm2.59}}$ & $8.62_{\pm0.16}$ \\
&dynamic &$\mathbf{54.5_{\pm1.34}}$ & $10.08_{\pm2.26}$ & $29.45_{\pm1.63}$& $\mathbf{49.83_{\pm2.25}}$ & $49.42_{\pm2.04}$ & $\mathbf{8.49_{\pm0.03}}$ \\
\midrule

\multirow{2}{*}{Heads}&static& $50.92_{\pm0.92}$ & $11.17_{\pm0.92}$ & $\mathbf{8.64_{\pm0.1}}$& $\mathbf{50.83_{\pm0.96}}$ & $\mathbf{51.42_{\pm1.01}}$ & $\mathbf{8.37_{\pm0.05}}$ \\
&dynamic  &$\mathbf{66.67_{\pm1.89}}$ & $\mathbf{20.58_{\pm2.46}}$ & $10.53_{\pm0.48}$ & $50.33_{\pm0.66}$ & $49.75_{\pm1.02}$ & $8.39_{\pm0.0}$ \\
\midrule

\multirow{2}{*}{Residual}&static&$\mathbf{81.33_{\pm2.96}}$ & $0.5_{\pm0.35}$ & $2465.4_{\pm158.71}$ & $\mathbf{77.83_{\pm1.74}}$ & $13.92_{\pm15.45}$ & $86.1_{\pm52.46}$ \\
&dynamic  &$59.25_{\pm0.94}$ & $\mathbf{13.92_{\pm1.96}}$ & $\mathbf{59.27_{\pm9.67}}$& $61.83_{\pm1.56}$ & $\mathbf{62.17_{\pm0.31}}$ & $\mathbf{8.52_{\pm0.02}}$ \\
\midrule

None&None &  $50.0_{\pm0.0}$ & $8.25_{\pm0.54}$ & $8.39_{\pm0.0}$ & $50.0_{\pm0.0}$ & $49.42_{\pm0.42}$ & $8.39_{\pm0.0}$ \\
\bottomrule

  \end{tabular}
  \label{Dynamic intervention vs. static intervention on DisentQA}
\end{table*}

\subsection{Finetuning facilitates intervention}\label{sec:pre_vs_finetune}

We explore the impact of fine-tuning on intervention success by comparing Llama2-7B vs.\ Goat-7B, a fine-tuned version of Llama2.
Table \ref{delta diff Dynamic intervention on the test set Llama2 vs. Goat} shows the average change resulting from intervention compared to no intervention (Full results are in Appendix \ref{appendix:finetune_vs_pre}). The intervention significantly enhances Goat's generation and classification accuracy compared to Llama2 across most components, without worsening perplexity. This suggests the advantage of utilizing the fine-tuned model over its basic counterpart, as shown in \cite{iti} but is not consistent across all studies \citep{truthx} and requires further research. %

\begin{table*}[t!]
  \caption{Llama2 and its fine-tuned model, Goat, showing differences from baseline (no intervention).}
\centering
\small
  
\begin{tabular}%
{p{0.08\linewidth}p{0.07\linewidth}p{0.1\linewidth}p{0.1\linewidth}p{0.1\linewidth}p{0.1\linewidth}p{0.1\linewidth}p{0.1\linewidth}}

\toprule 
 \multicolumn{2}{c}{Setting} & \multicolumn{3}{c}{DisentQA-\bn} & \multicolumn{3}{c}{TriviaQA-\bn}\\
\cmidrule(lr){1-2} \cmidrule(lr){3-5} \cmidrule(lr){6-8}

& & Classification & Generation & & Classification & Generation &  \\ 
& & \multicolumn{1}{c}{Accuracy} & \multicolumn{1}{c}{Accuracy}& Wiki ppl &  \multicolumn{1}{c}{Accuracy} &  \multicolumn{1}{c}{Accuracy} & Wiki ppl \\
\midrule

\multirow{2}{*}{Attention}&Llama2 &$15.0\hphantom{0}_{\pm1.02}$ & $12.0\hphantom{0}_{\pm1.67}$ & $\mathbf{2.65\hphantom{0}_{\pm0.51}}$  &$\hphantom{-}5.83\hphantom{0}_{\pm1.66}$ & $\hphantom{-}6.17\hphantom{0}_{\pm1.55}$ & $-0.01_{\pm0.02}$  \\
&Goat&$\mathbf{15.08_{\pm0.72}}$ & $\mathbf{15.33_{\pm1.12}}$ & $5.41\hphantom{0}_{\pm0.47}$& $\mathbf{\hphantom{-}9.08\hphantom{0}_{\pm3.8}}$ & $\mathbf{\hphantom{-}9.33\hphantom{0}_{\pm4.0}}$ & $\mathbf{\hphantom{-}0.0_{\pm0.0}}$\\\midrule

\multirow{2}{*}{MLP}&Llama2 &$4.5\hphantom{0}\hphantom{0}_{\pm1.34}$ & $1.83\hphantom{0}_{\pm1.84}$ & $21.06_{\pm1.63}$ &$-0.17\hphantom{0}_{\pm2.25}$ & $\hphantom{-}0.0\hphantom{0}\hphantom{0}_{\pm1.62}$ & $\hphantom{-}0.1_{\pm0.03}$ \\
&Goat &$\mathbf{8.92\hphantom{0}_{\pm2.55}}$ & $\mathbf{9.67\hphantom{0}_{\pm2.01}}$ & $\mathbf{11.46_{\pm1.34}}$ &$\mathbf{\hphantom{-}5.33\hphantom{0}_{\pm0.85}}$ & $\mathbf{\hphantom{-}6.17\hphantom{0}_{\pm0.12}}$ & $\hphantom{-}0.1_{\pm0.02}$  \\\midrule

\multirow{2}{*}{Heads}&Llama2  &$\mathbf{16.67_{\pm1.89}}$ & $12.33_{\pm1.94}$ & $\mathbf{2.14\hphantom{0}_{\pm0.48}}$&$\hphantom{-}0.33\hphantom{0}_{\pm0.66}$ & $\hphantom{-}0.33\hphantom{0}_{\pm0.62}$ & $\mathbf{\hphantom{-}0.0_{\pm0.0}}$ \\
&Goat&  $16.5\hphantom{0}_{\pm0.54}$ & $\mathbf{15.25_{\pm2.41}}$ & $3.47\hphantom{0}_{\pm0.29}$ & $\mathbf{\hphantom{-}1.17\hphantom{0}_{\pm2.93}}$ & $\mathbf{\hphantom{-}1.5\hphantom{0}\hphantom{0}_{\pm3.19}}$ & $\hphantom{-}0.01_{\pm0.01}$ \\\midrule

\multirow{2}{*}{Residual}&Llama2  &$9.25\hphantom{0}_{\pm0.94}$ & $5.67\hphantom{0}_{\pm1.43}$ & $50.89_{\pm9.67}$&  $\hphantom{-}11.83_{\pm1.56}$ & $\hphantom{-}12.75_{\pm0.54}$ & $\hphantom{-}0.14_{\pm0.02}$ \\
&Goat&$\mathbf{12.33_{\pm2.04}}$ & $\mathbf{9.67\hphantom{0}_{\pm3.63}}$ & $\mathbf{16.4\hphantom{0}_{\pm1.97}}$& $\mathbf{\hphantom{-}22.75_{\pm0.41}}$ & $\mathbf{\hphantom{-}23.58_{\pm1.03}}$ & $\mathbf{-0.05_{\pm0.15}}$  \\

\bottomrule
  \end{tabular}
  \label{delta diff Dynamic intervention on the test set Llama2 vs. Goat}
\end{table*}

\section{Related work}\label{sec:background}

Prior work has considered cases of LM failures under various terms such as factuality \citep{dola,LLM_Polygraph}, lying \citep{The_internal_state_of_an_llm_knows_when_its_lying,How_to_catch_an_ai_liar}, truthfulness \citep{geometry_of_truth,iti,truthx,Representation_engineering_lorra},  sycophancy \citep{Towards_understanding_sycophancy_in_language_models}, and hallucinations \citep{logits_context_modification,Chain-of-Verification}. 
Hallucinations are sometimes defined as cases of model mistakes that seem plausible to a user \citep{survey_of_hallucination_in_natural_language_generation}.

Research on hallucination typically falls into two main categories: detection and mitigation. Detecting hallucinations can involve treating the model as a black box, and posing questions or sampling \citep{gekhman-etal-2023-trueteacher,How_to_catch_an_ai_liar,SelfCheckGPT}. 
Another line of work has tried to detect hallucinations or factuality by examining the model's hidden  representations, often by training a detection classifier \citep{CCS, LLM_Polygraph, Weakly-Supervised_Detection_of_Hallucinations_in_LLM_Activations, The_Curious_Case_of_Hallucinatory_answerability, The_internal_state_of_an_llm_knows_when_its_lying, Do_Androids_Know_They're_Only_Dreaming, Constraint_Satisfaction_Lens_on_Factual_Errors, INSIDE, MLE_classification_using_inner_layers,levinstein2024still}
We similarly use detectors, emphasizing detection before a hallucination occurs.

Mitigating hallucinations can be done via prompting  \citep{Prompting_GPT-3_To_Be_Reliable, Chain-of-Verification, Chain_of_Natural_Language_Inference} or fine-tuning \citep{Fine-tuning_Language_Models_for_Factuality, Info_for_Faithfulness} or involve modifying the model's logits \citep{dola,Calibrated_language_models_must_hallucinate}.
Our work focuses on addressing hallucinations by intervening in the inner layers of the model \citep{truthx,iti,geometry_of_truth,Representation_engineering_lorra,trfr,nl-iti}, which potentially does not require external resources. 
One difference lies in closed-book versus open-book hallucinations  \citep{survey_of_hallucination_in_natural_language_generation,Do_Androids_Know_They're_Only_Dreaming}. Intervention body of work mostly studied the closed-book case, while we consider both cases, and characterize the effect of different choices for intervention, both from prior work and novel techniques we propose.

\section{Discussion and conclusion}

This work presented \bn, a framework for assessing white-box hallucination mitigation techniques in open-book and closed-book settings. We propose a typology of hallucinations through the lens of a model's knowledge. We highlight the possibility of hallucinations even though the model knows the answer, and propose an automatic way of creating a labeled dataset of such hallucinations. 

We then performed a comprehensive investigation of different variables involved in interventionist approaches for mitigating hallucinations.
 Notable insights include the importance of computing intervention vectors using pre-answer vectors instead of post-answer vectors; 
 showing that the novel dynamic intervention approach allows robust interventions; attention intervention consistently performs well across various measures and datasets, unlike the MLP component; classification accuracy is not always a good approximation for generation accuracy and perplexity should be calculated as well.

This study provides a comprehensive framework for creating knowledge-based hallucination datasets and offers insights into effective intervention strategies in both open-book and closed-book settings.

\section{Limitations and ethics statement}\label{sec:limitations}
This work considers a wide range of configurations for mitigating hallucinations.
One limitation is that we only focused on one model (Llama2-7B) and one of its fine-tuned variants (Goat-7B). Additionally, our results span one open-book dataset (DisentQA) and one closed-book dataset (TriviaQA).
Lastly, we use an initial setting and modify a single variable each time instead of checking across all options. To form a broader understanding, additional settings and models must be investigated.

The primary objective of our work is to create a better understanding of how to mitigate hallucinations. By utilizing interventions, a malicious actor could intervene in the opposite direction and increase hallucinations. However, our research aims to increase understanding and mitigation of hallucinations.

\section{Acknowledgement}
This research has been supported by an AI Alignment grant from Open Philanthropy, the Israel Science Foundation (grant No. 448/20), the Azrieli Foundation Early Career Faculty Fellowship, and by a grant under a Master Sponsored Research Agreement between the Technion and Google. We also thank Google Cloud for providing us with credits for running experiments on the Google Cloud Platform.

\bibliography{references}

\appendix
\section{Dataset construction specifics}\label{appendix:Dataset_generation_specifics}

\subsection{General dataset construction specifics}\label{appendix:General Dataset Construction Specifics}
We created at least 1,000 examples labeled as ``hallucination" and 1,000 as ``grounded". We filtered out instances where the golden/parametric/contextual answers exceeded 5 tokens or where the prompt exceeded 1,000 tokens.

To assess the model's knowledge of the golden answer, we employed a beam search generation of 20 tokens, repeated 5 times with a temperature of 0.8. We utilized only examples falling under type 3 knowledge (Section \ref{sec:Hallucination Example Selection}), ensuring each answer was generated at least four times.

\subsection{Open-book dataset construction specifics}\label{appendix:Open-Book Dataset Construction Specifics}
We loaded the DisentQA (open-book) dataset from \url{https://docs.google.com/document/d/1Z4vA7ifMQTk5YBF3BEYCFSnIvXCYaznLP_7VBcXPEeU/edit#heading=h.3prdi119z3tn} and used <P> type counterfactual examples from the dataset with the name \emph{v10-simplified\_simplified-nq-train\_factual\_counterfactual\_disentangled\_baseline\_train\_split.csv}.

In the open book setting, to determine the model's rank in generating the parametric/contextual answer, we computed the difference in rank of the first token that differs in the answer tokens between the two answers.
And used 1 as the threshold $T$.

\subsection{Closed-book dataset construction specifics}\label{appendix:Closed-Book Dataset Construction Specifics}
In the closed-book setting, the rank was determined using the average rank of all answer tokens. Additionally, given the extensive size of TriviaQA, we randomly sampled 20,000 examples to construct our labeled dataset and loaded it from the "datasets" library. Table \ref{Generated answers using bad shots in the prompt} shows examples of generated hallucinations of Llama2 using bad-shots in greedy decoding.

\begin{table*}[t!]

\caption{Generated answers using greedy decoding on Llama2 on a prompt with bad shots in the beginning with the contamination of a question from (closed-book) TrivaQA.}
\centering
\small
  
  \begin{tabular}  
  {p{0.55\linewidth}p{0.1\linewidth}p{0.1\linewidth}}

\multicolumn{1}{c}{Prompt} &\multicolumn{1}{c}{Grounded} & \multicolumn{1}{c}{Hallucination}\\\midrule
question: What is the capital of France?\textbackslash nwrong answer: Berlin\textbackslash nquestion: Who painted the Mona Lisa?\textbackslash nwrong answer: Pablo Picasso\textbackslash nquestion: What is the powerhouse of the cell?\textbackslash nwrong answer: Golgi Apparatus\textbackslash nquestion: Which English king is buried in Worcester Cathedral?\textbackslash nwrong answer:&King John & Henry VIII \\\midrule
question: Who wrote 'To Kill a Mockingbird'?\textbackslash nwrong answer: J.K. Rowling\textbackslash nquestion: What is the smallest prime number?\textbackslash nwrong answer: 1\textbackslash nquestion: Who discovered penicillin?\textbackslash nwrong answer: Isaac Newton\textbackslash nquestion: Which colour lies between red and yellow in the visible spectrum\textbackslash nwrong answer:&Orange& Green
\\\midrule
question: Who wrote 'To Kill a Mockingbird'?\textbackslash nwrong answer: J.K. Rowling\textbackslash nquestion: What is the chemical symbol for gold?\textbackslash nwrong answer: Ag\textbackslash nquestion: Which element has the chemical symbol 'H'?\textbackslash nwrong answer: Helium\textbackslash nquestion: In Greek Mythology, who was the Mother of Andromeda?\textbackslash nwrong answer:&Cassiopeia& Andromeda

  \\\bottomrule

  \end{tabular}
  
  \label{Generated answers using bad shots in the prompt}
\end{table*}

Lastly, the full list of the 20 bad shots and good shots:

\textbf{Good shots:}

question: What is the capital of France?\\answer: Paris\\question: How many continents are there?\\answer: 7\\question: Who wrote 'Romeo and Juliet'?\\answer: William Shakespeare\\question: What is the square root of 64?\\answer: 8\\question: Which element has the chemical symbol 'H'?\\answer: Hydrogen\\question: Who was the first President of the United States?\\answer: George Washington\\question: What is the powerhouse of the cell?\\answer: Mitochondria\\question: In what year did World War II end?\\answer: 1945\\question: What is the currency of Japan?\\answer: Japanese Yen\\question: Who painted the Mona Lisa?\\answer: Leonardo da Vinci\\question: What is the speed of light?\\answer: 299,792 kilometers per second\\question: How many sides does a hexagon have?\\answer: 6\\question: What is the boiling point of water in Celsius?\\answer: 100 degrees\\question: Who wrote 'To Kill a Mockingbird'?\\answer: Harper Lee\\question: What is the capital of Australia?\\answer: Canberra\\question: What is the largest ocean on Earth?\\answer: Pacific Ocean\\question: Who discovered penicillin?\\answer: Alexander Fleming\\question: What is the chemical symbol for gold?\\answer: Au\\question: What is the smallest prime number?\\answer: 2\\question: How many planets are there in our solar system?\\answer: 8\\

\textbf{Bad shots:}

question: What is the capital of France?\\answer: Berlin\\question: How many continents are there?\\answer: 6\\question: Who wrote 'Romeo and Juliet'?\\answer: Jane Austen\\question: What is the square root of 64?\\answer: 7\\question: Which element has the chemical symbol 'H'?\\answer: Helium\\question: Who was the first President of the United States?\\answer: Abraham Lincoln\\question: What is the powerhouse of the cell?\\answer: Golgi Apparatus\\question: In what year did World War II end?\\answer: 1939\\question: What is the currency of Japan?\\answer: Euro\\question: Who painted the Mona Lisa?\\answer: Pablo Picasso\\question: What is the speed of light?\\answer: 300,000 kilometers per second\\question: How many sides does a hexagon have?\\answer: 5\\question: What is the boiling point of water in Celsius?\\answer: 50 degrees\\question: Who wrote 'To Kill a Mockingbird'?\\answer: J.K. Rowling\\question: What is the capital of Australia?\\answer: Sydney\\question: What is the largest ocean on Earth?\\answer: Atlantic Ocean\\question: Who discovered penicillin?\\answer: Isaac Newton\\question: What is the chemical symbol for gold?\\answer: Ag\\question: What is the smallest prime number?\\answer: 1\\question: How many planets are there in our solar system?\\answer: 9\\

\section{Final datasets statistic}\label{appendix:Datasets Statistic}
In section \ref{sec:dataset_creation} we created a labeled dataset given a model and a dataset. DisentQA for the open-book dataset and TriviaQA for the closed-book dataset. The statistic of the final dataset count of the number of examples in each label can be found in Table \ref{Dataset_Statistic}, the else label is all the examples that did not classify to the other two labels.
We can see that the different settings have different results, indicating the importance of creating a labeled dataset regarding the specific model.

\begin{table*}[t!]

  \caption{Dataset label statistics.}

\centering
\small
  
  \begin{tabular}{llllp{0.1\linewidth}p{0.1\linewidth}p{0.1\linewidth}p{0.1\linewidth}p{0.1\linewidth}p{0.1\linewidth}p{0.1\linewidth}}
  \multicolumn{1}{c}{Dataset name} &\multicolumn{1}{c}{ \# Grounded} & \multicolumn{1}{c}{\# Hallucination}& \multicolumn{1}{c}{\# Else}\\\midrule

closed-book-Llama2-TriviaQA-\bn&4553&1451&10306\\
open-book-Llama2-DisentQA-\bn&1673&2869&8958\\
closed-book-Goat-TriviaQA-\bn&2618&1507&12185\\
open-book-Goat-DisentQA-\bn&3019&1261&9220\\

\bottomrule

  \end{tabular}
  \label{Dataset_Statistic}
\end{table*}

\section{Specific evaluation measures}\label{Specific_Evaluation_Measures+On_no_context_setting}

In the open book setting, we assess classification accuracy by comparing the model's preference for the \contextual,  Rank(\contextual)$\le$Rank(\parametric)
 Regarding the closed book setting, we compare Rank(\golden | bad shots) $\le$ Rank(\golden | good shots).
In both settings, the rank calculation mirrors that of the dataset creation step. If the equations are true the example is labeled `grounded', else `hallucination'.

Generation accuracy in the open book setting evaluates whether the \contextual was generated and the \parametric was not.
In the closed book setting evaluates whether the golden answer \golden is produced.

Lastly, to calculate the Wiki ppl we use 100 random articles, and from each, we use the first 100 tokens to calculate the perplexity.
\section{Detection results}\label{appendix:Detection results}

\begin{figure*}
\centering
\begin{subfigure}{0.48\textwidth}
  \centering
  \includegraphics[width=\linewidth]{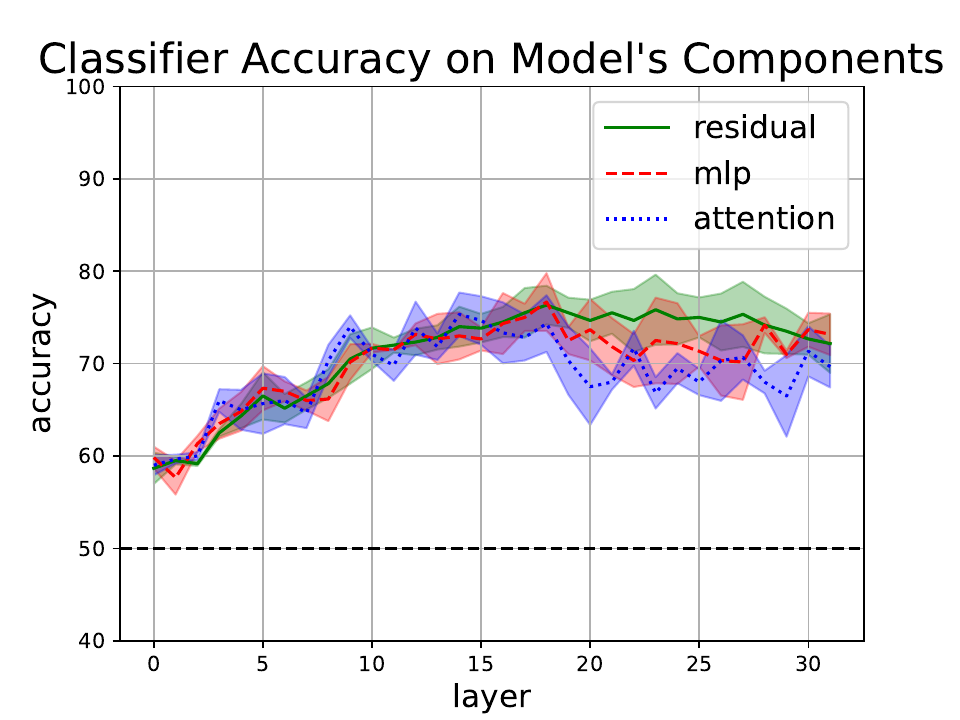}
  \caption{DisentQA-\bn (open book) dataset.}
  \label{fig:sub1_main}
 \end{subfigure}%
 \hfill
 \begin{subfigure}{0.48\textwidth}
  \centering
  \includegraphics[width=\linewidth]{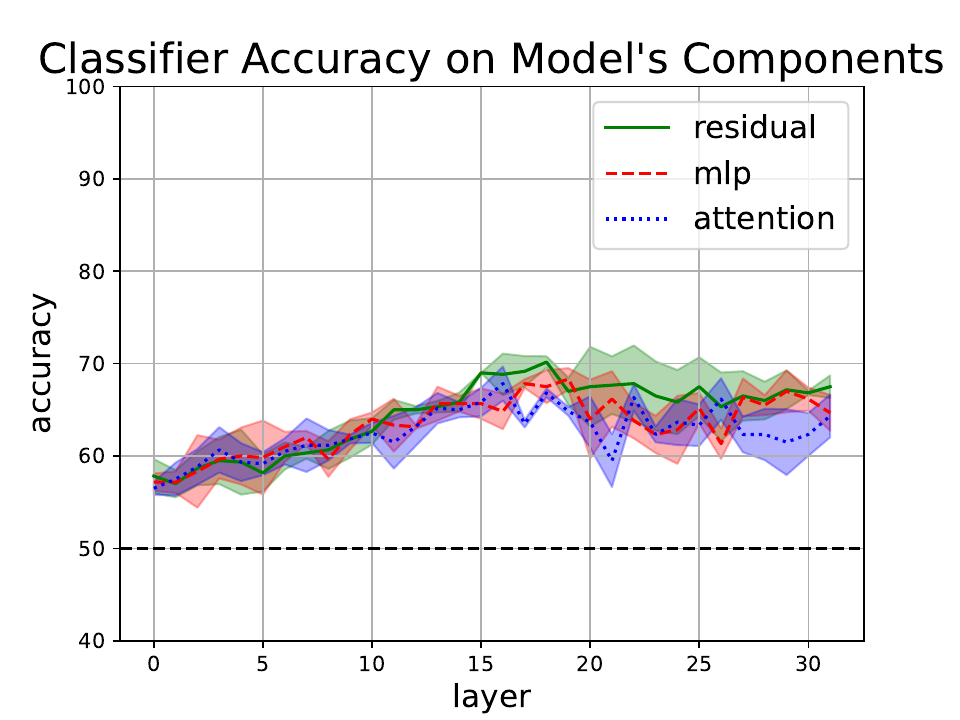}
  \caption{TriviaQA-\bn (closed-book) dataset.}
  \label{fig:sub2_main}
 \end{subfigure}
 
\caption{Hallucination detection accuracy using different layers and components of Llama2-7B. Detection is much better than random, but still far from perfect.}
\label{figm}
\end{figure*}

Figures \ref{fig:sub1_main} and \ref{fig:sub2_main} show hallucination detection results on the (open-book) DisentQA-\bn and (closed-book) TriviaQA-\bn, respectively, when applied to different model components at different layers, using Llama-2-7B. 

On most layers, detection is significantly higher than random (dashed lines), indicating a signal of possible hallucination \emph{before} the occurrence of hallucination. Additionally, we observe that the different components yield fairly similar results, with a slight advantage for detection based on the residual stream.

\begin{figure*}
\centering
\begin{subfigure}{0.48\textwidth}
  \centering
  \includegraphics[width=\linewidth]{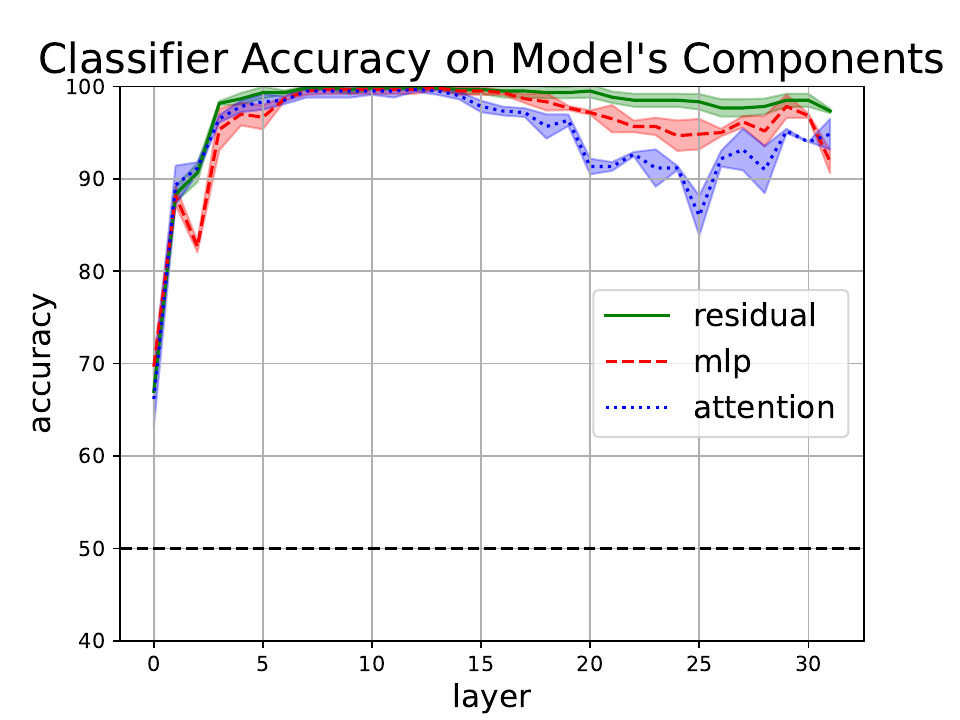}
  \caption{}
 \end{subfigure}%
 \hfill
 \begin{subfigure}{0.48\textwidth}
  \centering
  \includegraphics[width=\linewidth]{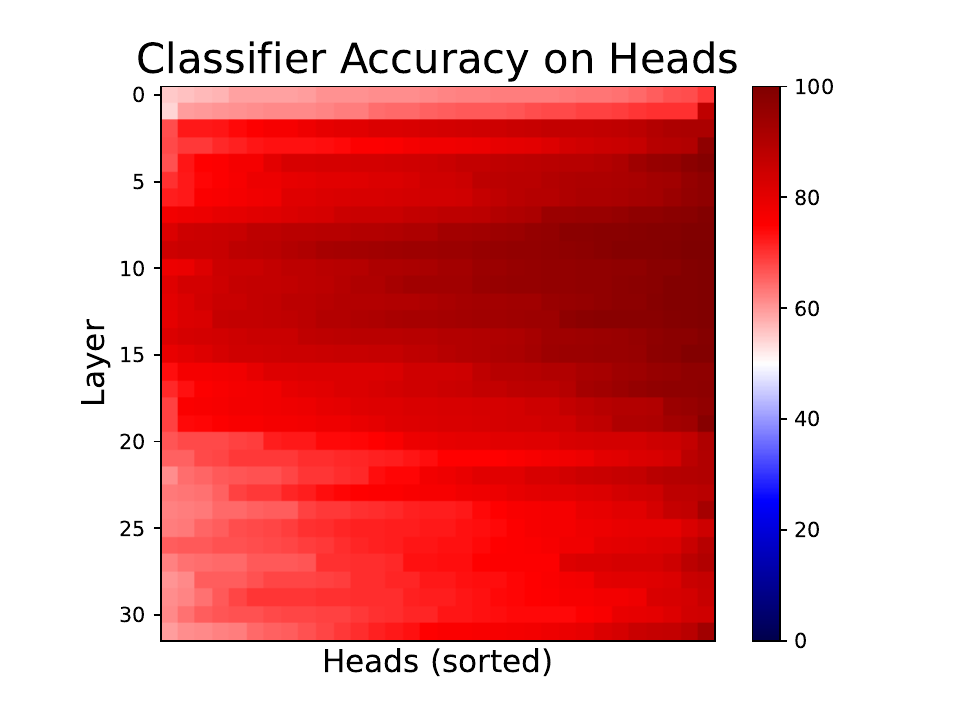}
  \caption{}
 \end{subfigure}
 
\caption{Accuracy in detecting hallucination post answer on Llama2 DisentQA-\bn (open book-setting).}
\label{Accuracy in detecting hallucination post answer on Llama2 DisentQA (open book-setting).}
\end{figure*}

We remind the reader that detecting possible hallucinations before they occur is a more challenging task than detecting them post-hallucination \citep{Do_Androids_Know_They're_Only_Dreaming}. Indeed, if we concatenate the answer to the prompt, we reach over 90\% accuracy in most layers and components, see Figure \ref{Accuracy in detecting hallucination post answer on Llama2 DisentQA (open book-setting).}.\footnote{We run this test only on the open-book setting as it has a clear answer to concatenate (\contextual for grounded examples and \parametric for hallucination examples).} Given that in practical use cases the answer is not known ahead of generation, we emphasize the importance of pre-hallucination detection.

\begin{figure*}
\centering
\begin{subfigure}{0.33\textwidth}
  \centering
  \includegraphics[width=\linewidth]{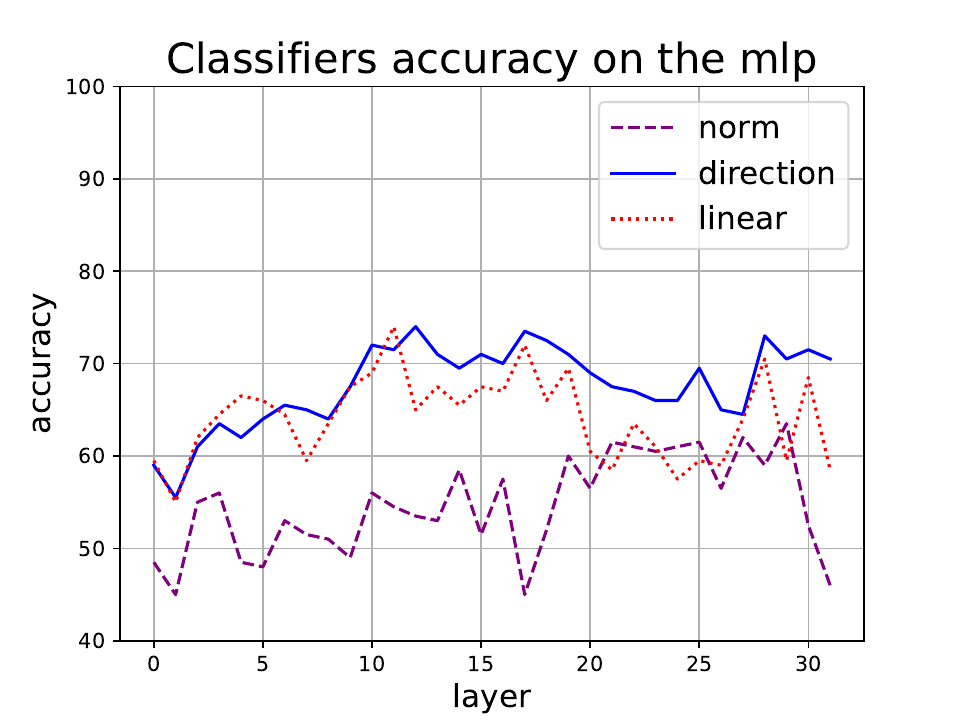}
  \caption{MLP}
  \label{fig:sub1_mlp}
 \end{subfigure}%
 \hfill
 \begin{subfigure}{0.33\textwidth}
  \centering
  \includegraphics[width=\linewidth]{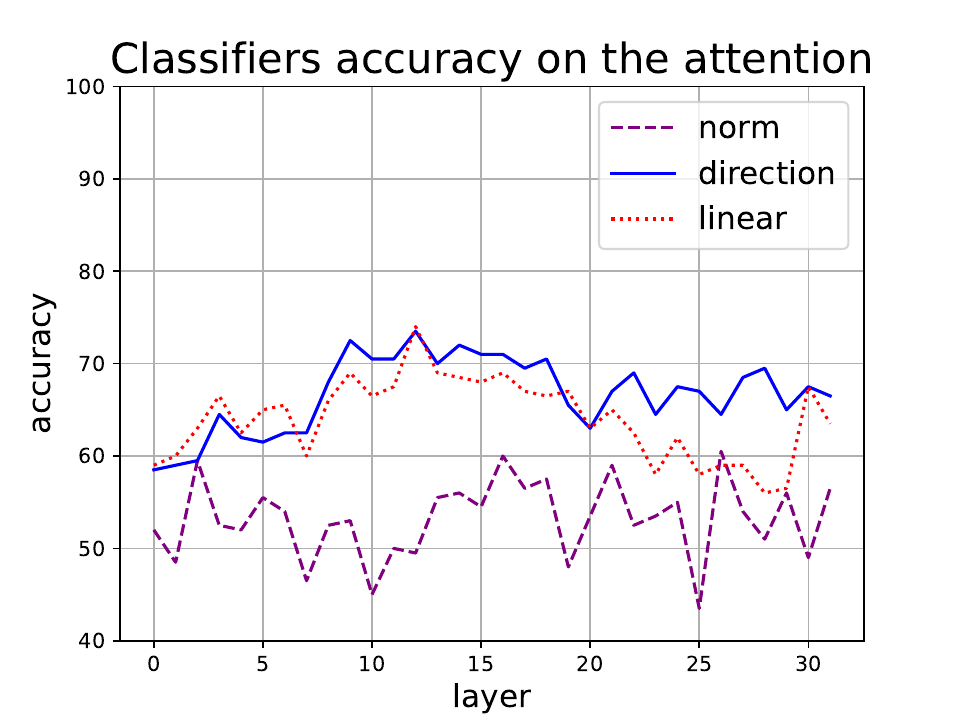}
  \caption{Attention}
  \label{fig:sub2_attn}
 \end{subfigure}
 \hfill
 \begin{subfigure}{0.33\textwidth}
  \centering
  \includegraphics[width=\linewidth]{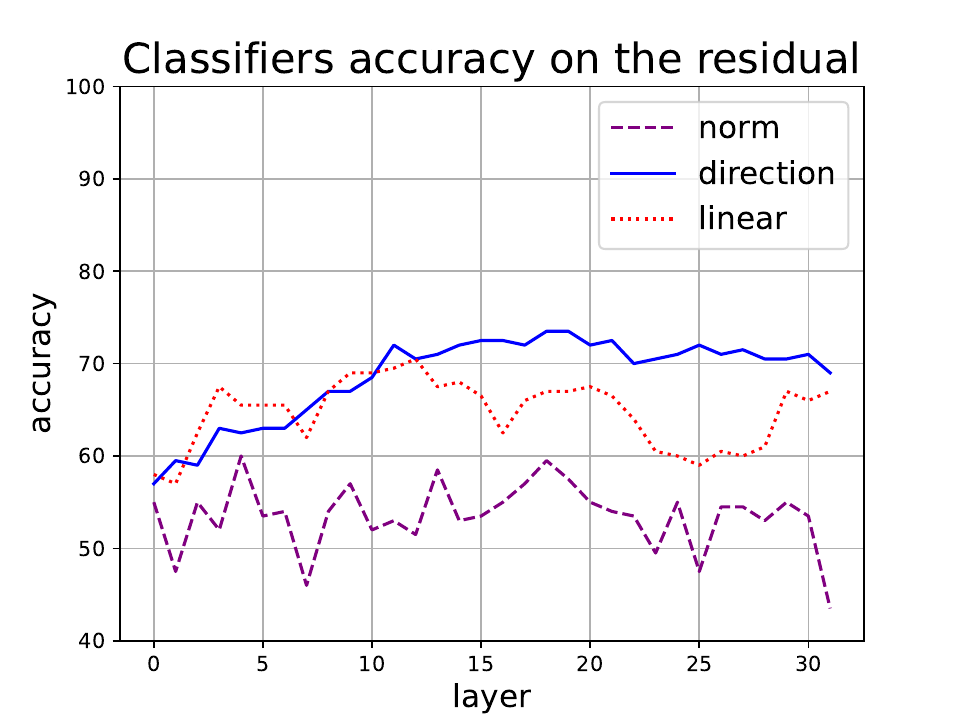}
  \caption{Residual}
  \label{fig:sub3_aresidual}
 \end{subfigure}
 
\caption{Accuracy in detecting hallucination in \emph{open book setting} of DisentQA-\bn using three classifiers.}
\label{fig_mlp_attn}
\end{figure*}

\begin{figure*}
\centering

 \begin{subfigure}{0.33\textwidth}
     \includegraphics[width=\textwidth]{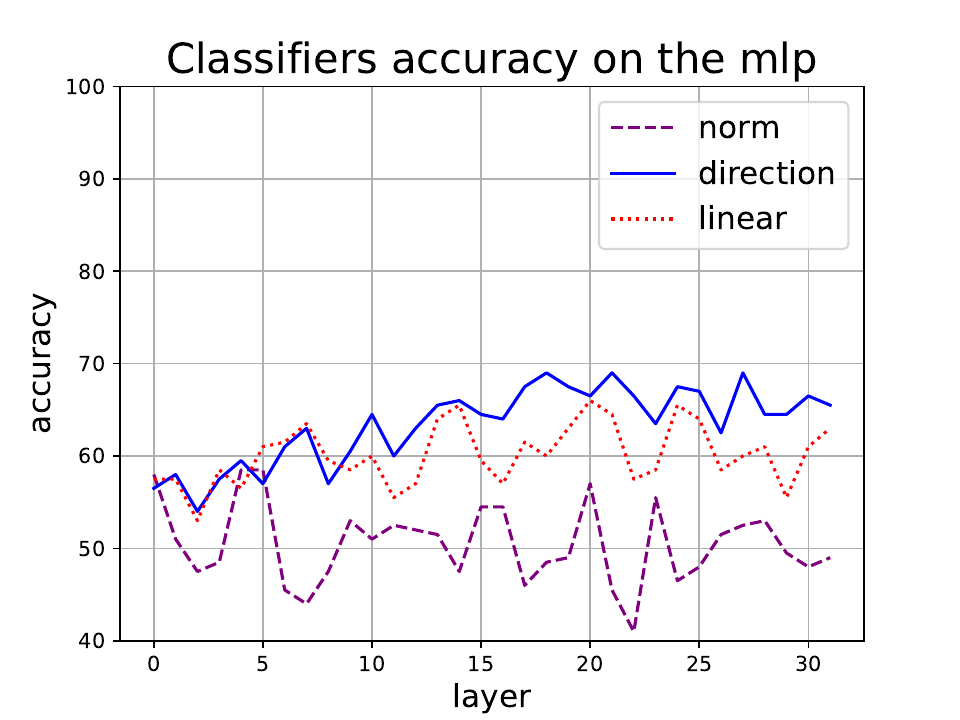}
     \caption{MLP}
     \label{fig:a}
 \end{subfigure}%
 \hfill
 \begin{subfigure}{0.33\textwidth}
     \includegraphics[width=\textwidth]{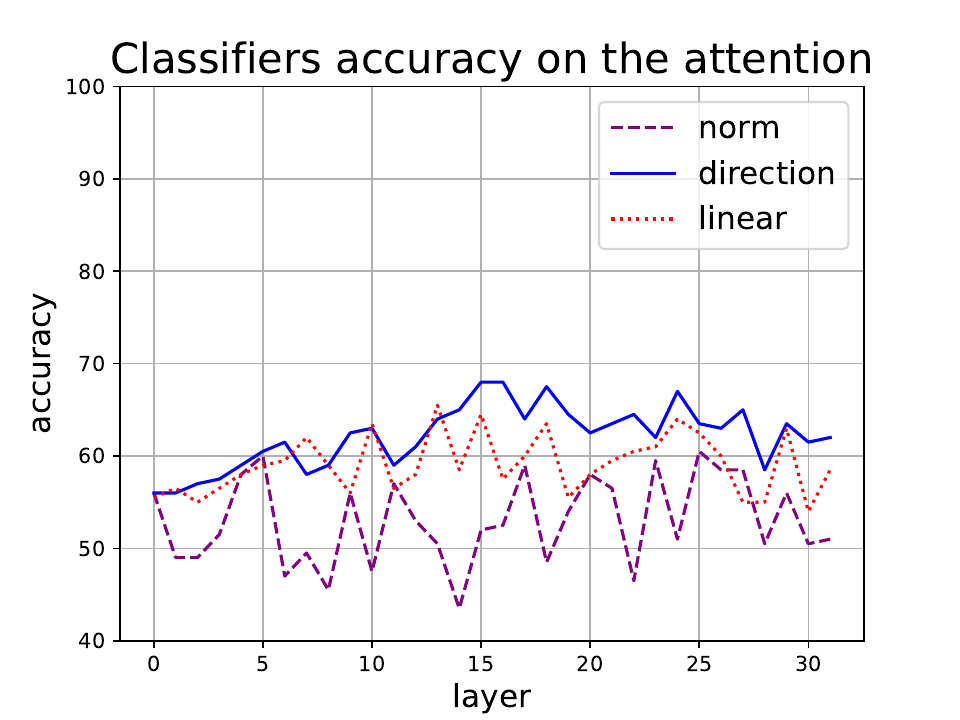}
     \caption{Attention}
     \label{fig:b}
 \end{subfigure}%
 \hfill 
  \begin{subfigure}{0.33\textwidth}
  \centering
  \includegraphics[width=\linewidth]{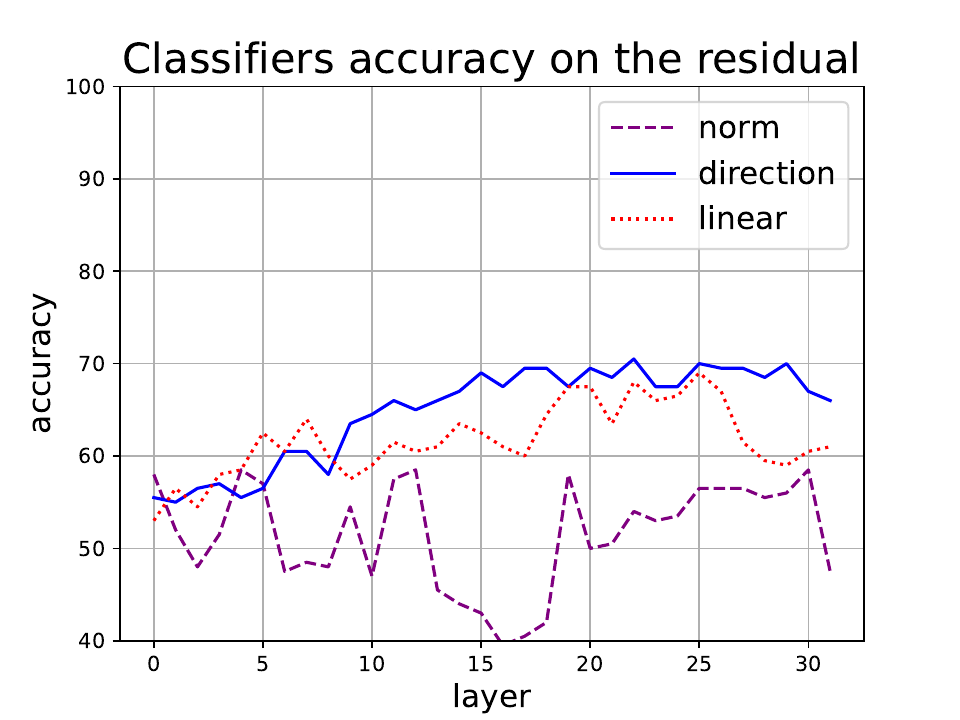}
  \caption{Residual}
  \label{fig:c}
 \end{subfigure}\textbf{}

 \caption{Accuracy in detecting hallucination in \emph{closed book setting} of TriviaQA-\bn using three classifiers.}
 \label{fig_no_context_detection}

\end{figure*}

For completeness, we present the detection results on one data split of the validation set on the residual, attention, and MLP layers in Figures \ref{fig_mlp_attn} and \ref{fig_no_context_detection}, utilizing three different classifiers: a linear classifier trained on the vectors as they are (labeled classifier in the figures), a linear classifier trained only on the vectors' norms (labeled norm in the figures), and a linear classifier trained on the normalized vectors (labeled direction in the figures). It's evident that the linear classifier trained on the normalized vectors (direction classifier) achieves the best results across most components, layers, and datasets. Thus we use normalized vectors for the detection in all results.

\begin{figure*}
\centering
\begin{subfigure}{0.48\textwidth}
  \centering
  \includegraphics[width=\linewidth]{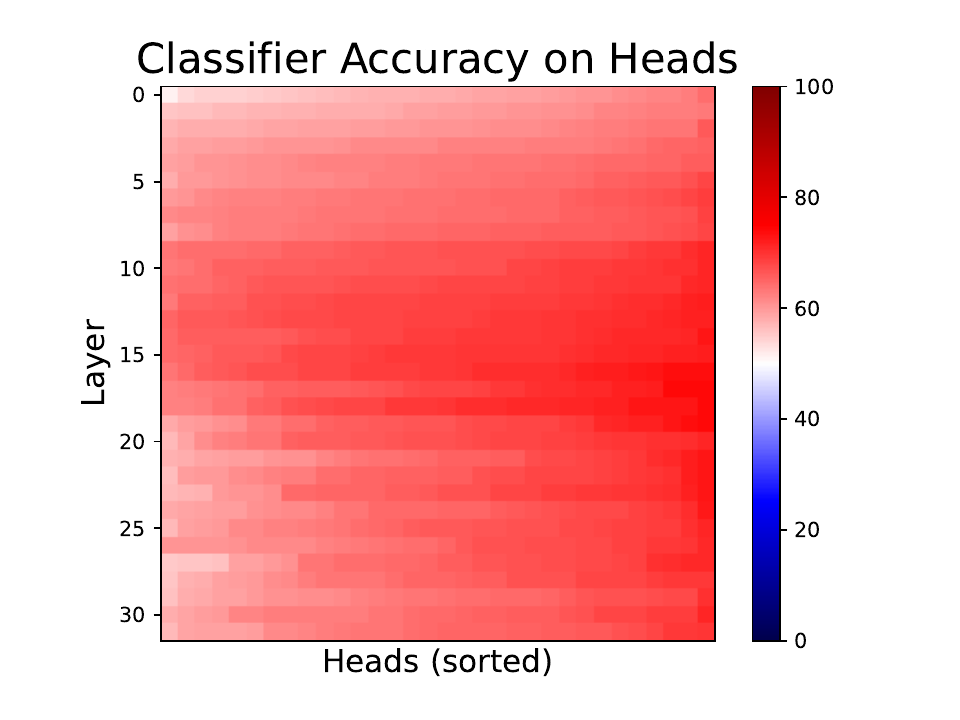}
  \caption{DisentQA-\bn (open book) dataset.}
  \label{fig:disentqa dataset heads}
 \end{subfigure}%
 \hfill
 \begin{subfigure}{0.48\textwidth}
  \centering
  \includegraphics[width=\linewidth]{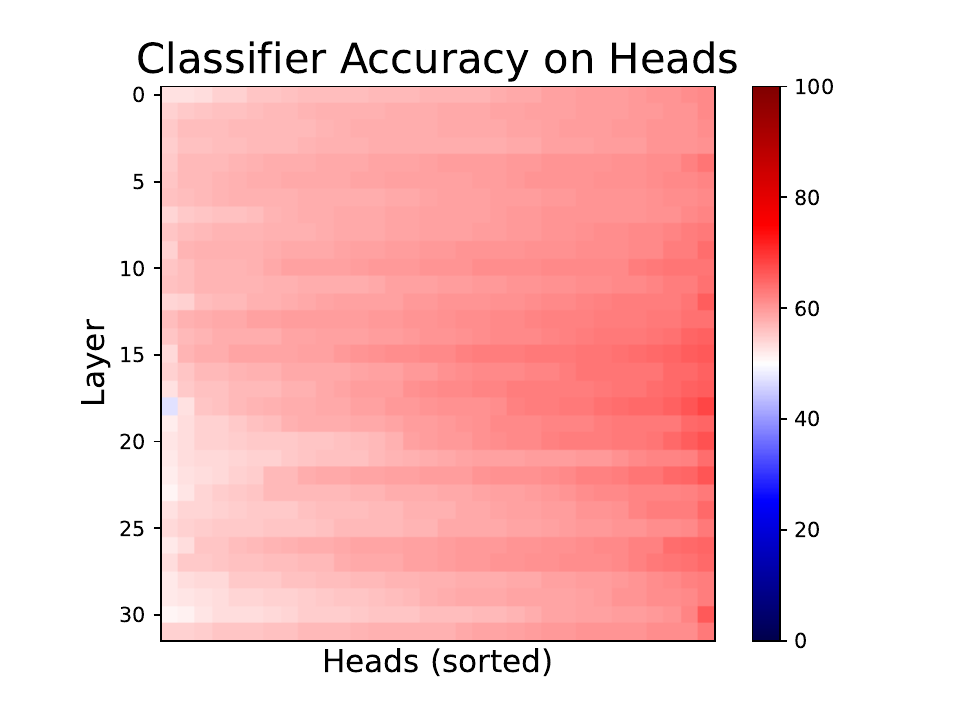}
  \caption{TriviaQA-\bn (closed book) dataset.}
  \label{fig:TriviaQA dataset heads}
 \end{subfigure}
 
\caption{Accuracy in detecting hallucination on Llama2 heads.}
\label{appendix:Accuracy in detecting hallucination on Llama2 heads.}
\end{figure*}

Additionally in Figure \ref{appendix:Accuracy in detecting hallucination on Llama2 heads.} we see the classification results on the heads on Llama2-7B in the two settings. We can see that those results are a bit less good than the ones on the other model's components with much better results in the open-book setting than the closed-book setting.

\begin{figure*}
\centering
\begin{subfigure}{0.48\textwidth}
  \centering
  \includegraphics[width=\linewidth]{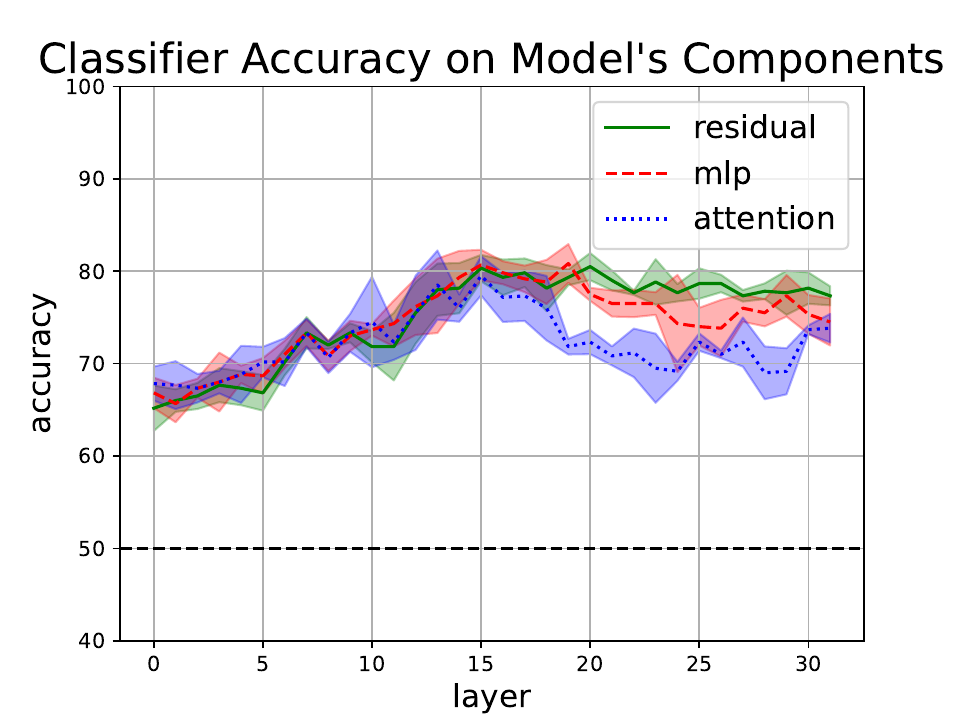}
  \caption{DisentQA-\bn dataset.}
  \label{fig:llama_classifier_few_seeds}
 \end{subfigure}%
 \hfill
 \begin{subfigure}{0.48\textwidth}
  \centering
  \includegraphics[width=\linewidth]{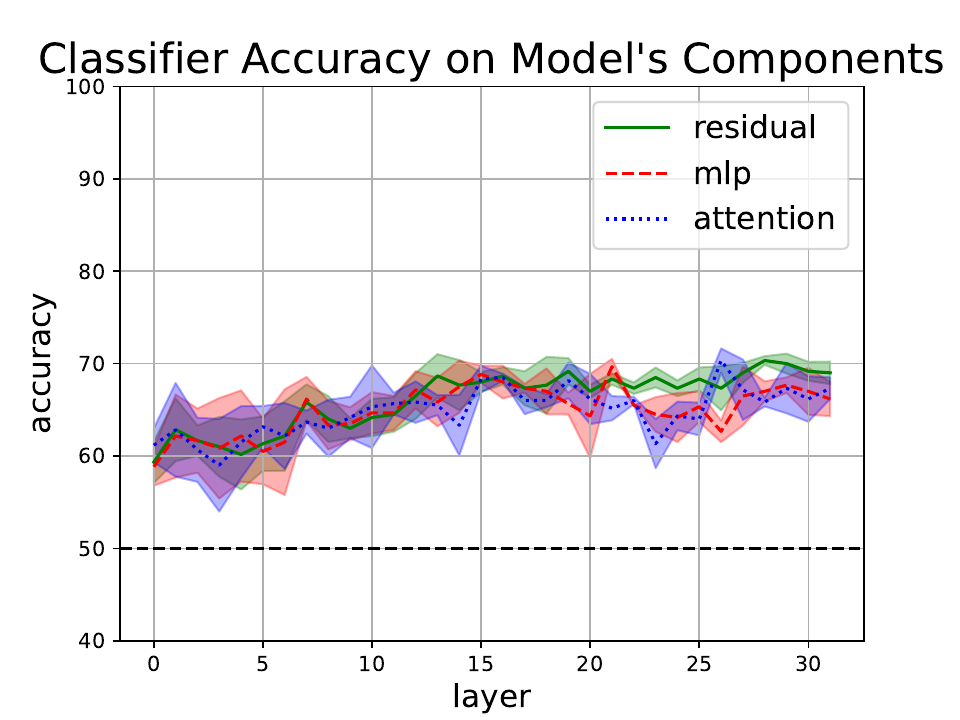}
  \caption{TriviaQA-\bn dataset.}
  \label{fig:goat_classifier_few_seeds}
 \end{subfigure}
 
\caption{Accuracy in detecting hallucination across the layers and components of Goat-7B model.}
\label{appendix:fig_goat_vs_detection_classifier}
\end{figure*}

\begin{figure*}
\centering
\begin{subfigure}{0.48\textwidth}
  \centering
  \includegraphics[width=\linewidth]{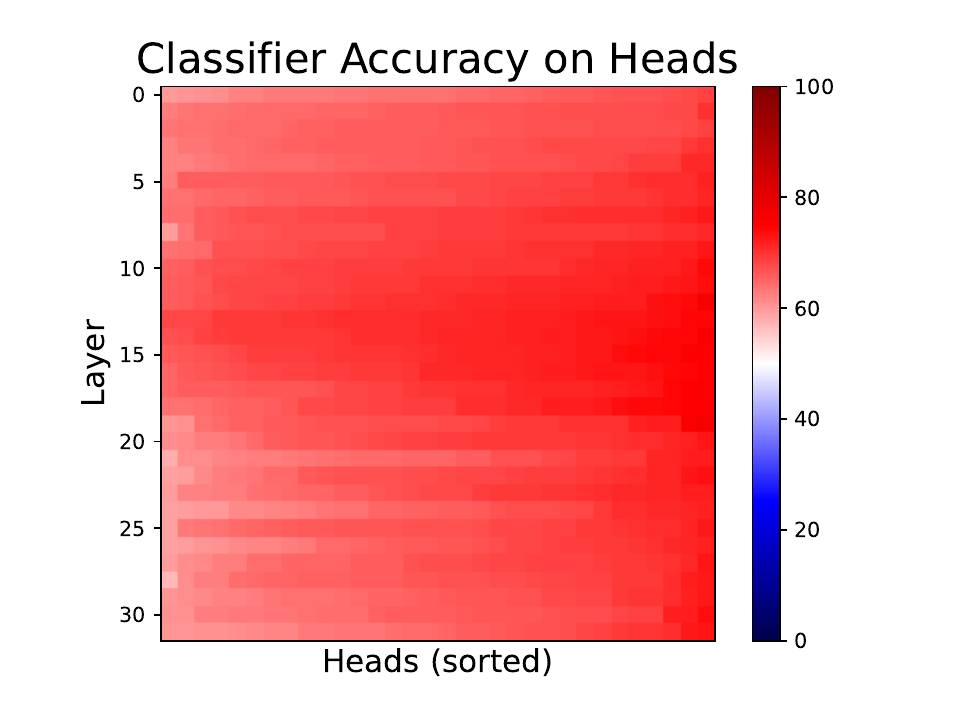}
  \caption{DisentQA-\bn (open book) dataset.}
  \label{fig:disentqa dataset heads goat}
 \end{subfigure}%
 \hfill
 \begin{subfigure}{0.48\textwidth}
  \centering
  \includegraphics[width=\linewidth]{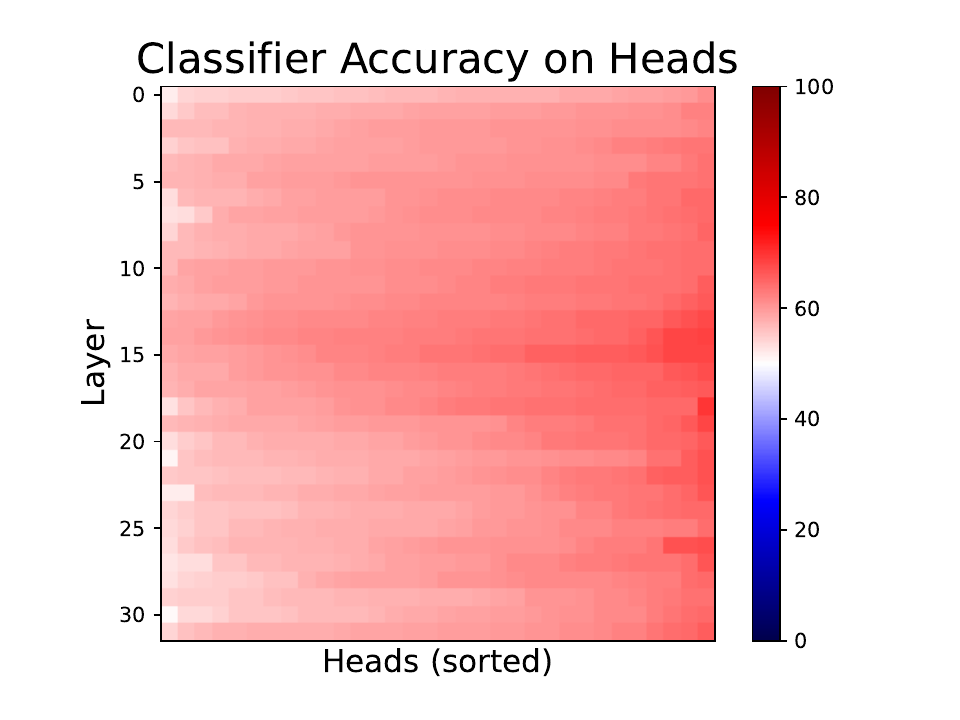}
  \caption{TriviaQA-\bn (closed book) dataset.}
  \label{fig:TriviaQA dataset heads goat}
 \end{subfigure}
 
\caption{Accuracy in detecting hallucination on Goat heads.}
\label{appendix:Accuracy in detecting hallucination on Goat heads.}
\end{figure*}

Lastly, in Figures \ref{appendix:fig_goat_vs_detection_classifier} and \ref{appendix:Accuracy in detecting hallucination on Goat heads.}, we observe the direction classifier's accuracy across the Goat-7B model's layers on both datasets. The detection results are similar and slightly better than those on Llama2.

\begin{figure*}
\centering
\begin{subfigure}{0.48\textwidth}
  \centering
  \includegraphics[width=\linewidth]{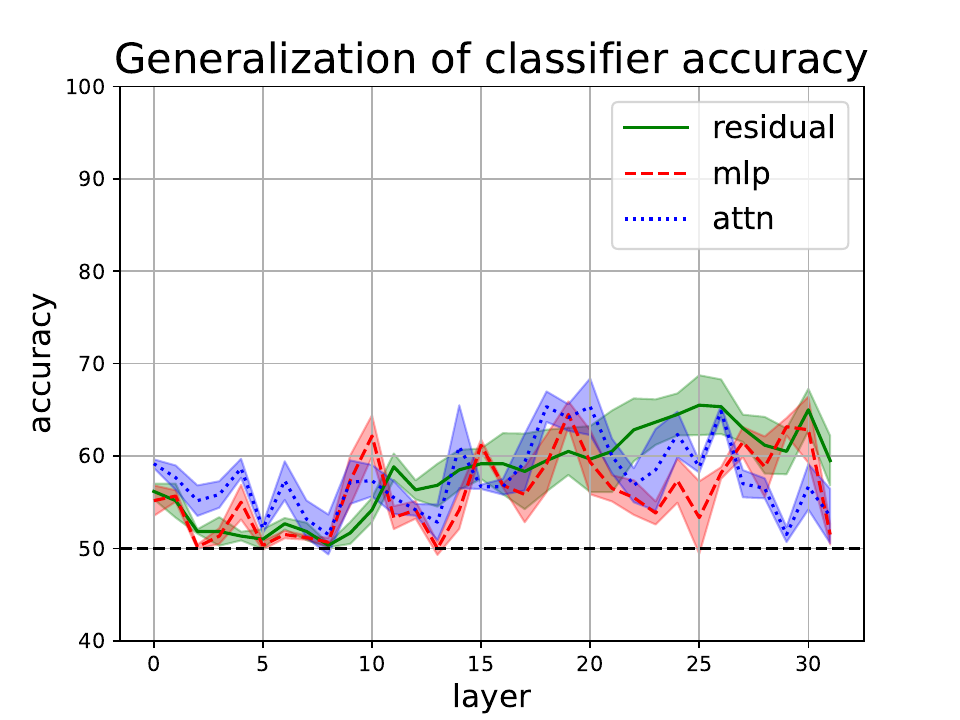}
  \caption{Classifier trained on Goat-7B using DisentQA-\bn dataset.}
  \label{fig:Classifier trained on Goat-7B using DisentQA dataset}
 \end{subfigure}%
 \hfill
 \begin{subfigure}{0.48\textwidth}
  \centering
  \includegraphics[width=\linewidth]{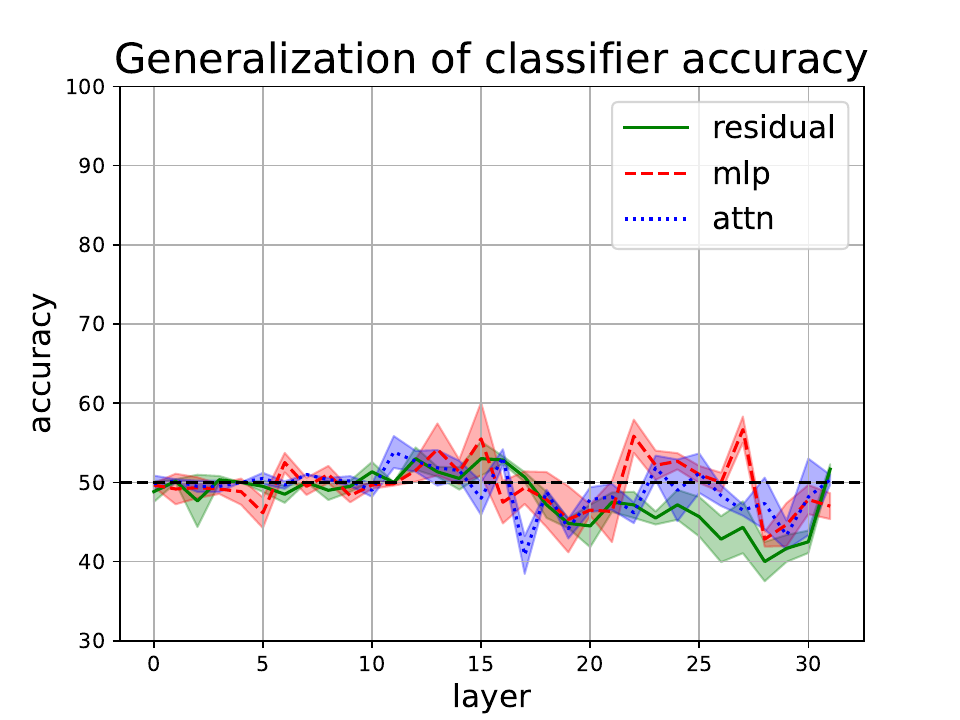}
  \caption{Classifier trained on Llama2-7B using TriviaQA-\bn dataset.}
  \label{fig:Classifier trained on Llama2-7B using TriviaQA dataset}
 \end{subfigure}
 
\caption{Accuracy in detecting hallucination across the layers of Llama2-7B and DisentQA-\bn (open-book dataset).}
\label{fig_goat_vs_llama_classifier}
\end{figure*}

So far we have performed each experiment within a given dataset and model. 
Do hallucination detectors generalize across datasets or models? 
 In Figure \ref{fig:Classifier trained on Goat-7B using DisentQA dataset}, we employ a classifier trained on Goat-7B and evaluate its performance when detecting hallucinations from Llama2-7B states, on the DisentQA-\bn dataset (Goat$\rightarrow$Llama2; DisentQA-\bn).  %
 Interestingly, the Goat classifier consistently outperforms random chance across all layers, suggesting a notable correlation between the inner states of the two models. These results correspond to related works \citep{wu2020similarity}.
 However, the results are much lower than a detector trained and evaluated on Llama2 states (Llama2$\rightarrow$Llama2;  \ref{fig:sub1_main}), indicating that cross-model generalization is limited.

In contrast, Figure \ref{fig:Classifier trained on Llama2-7B using TriviaQA dataset} illustrates that detectors struggle to generalize across the two datasets. This underscores that to help understand and mitigate hallucinations it is important to address each hallucination scenario individually and make cautious claims about each one. 

\begin{figure*}
\centering
\begin{subfigure}{0.48\textwidth}
  \centering
  \includegraphics[width=\linewidth]{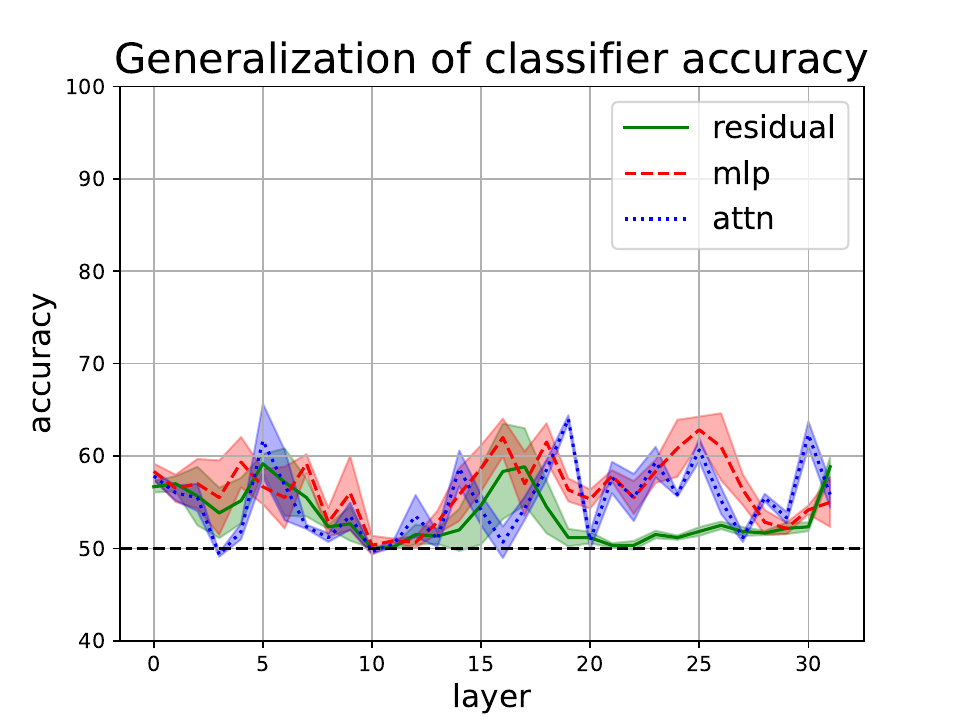}
  \caption{Classifier trained on Goat-7B using TriviaQA-\bn dataset.}
  \label{fig_appendix:goat_classifier}
 \end{subfigure}%
 \hfill
 \begin{subfigure}{0.48\textwidth}
  \centering
  \includegraphics[width=\linewidth]{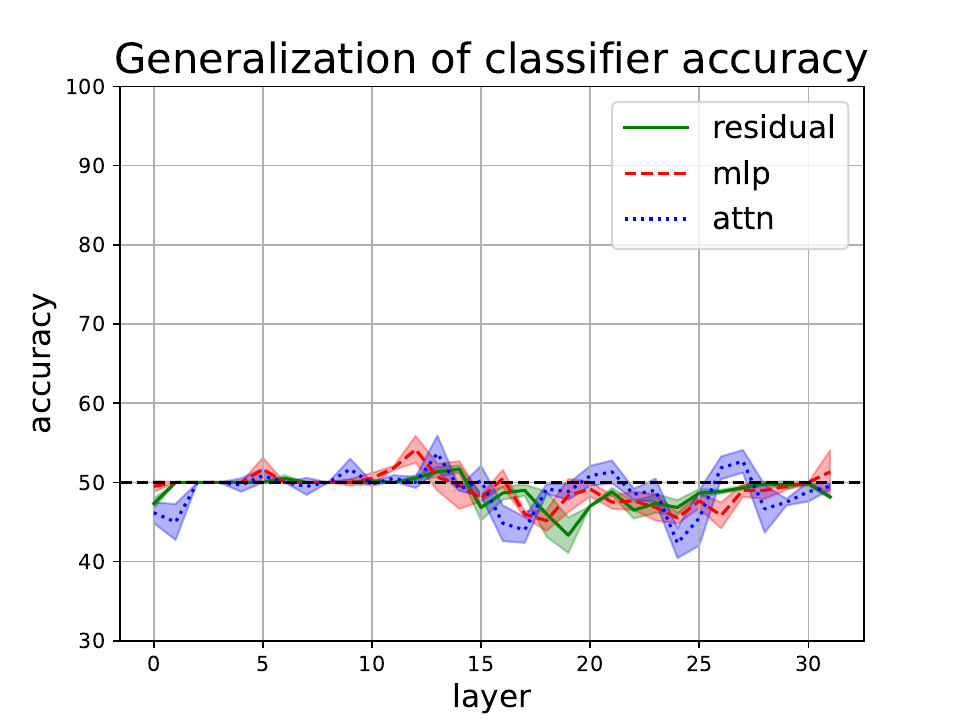}
  \caption{Classifier trained on Llama2 using DisentQA-\bn dataset.}
  \label{fig_appendix:disentqa_detect_triviaqa}
 \end{subfigure}
 
\caption{Accuracy in detecting hallucination across the layers of Llama2 and TriviaQA-\bn (closed-book dataset).}
\label{fig_goat_vs_llama_classifier_trivia_qa}
\end{figure*}

\begin{figure*}
\centering
\begin{subfigure}{0.48\textwidth}
  \centering
  \includegraphics[width=\linewidth]{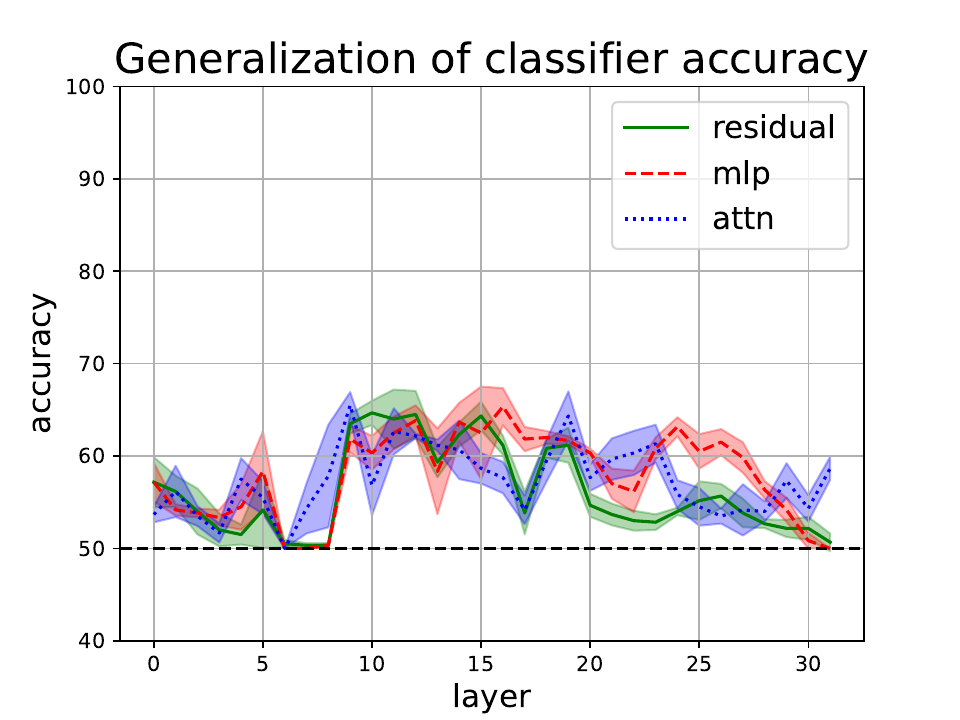}
  \caption{Classifier trained on Llama2 using TriviaQA-\bn and tested on Goat using TriviaQA-\bn.}
  \label{fig_appendix:trivia_detect_disent}
 \end{subfigure}%
 \hfill
 \begin{subfigure}{0.48\textwidth}
  \centering
  \includegraphics[width=\linewidth]{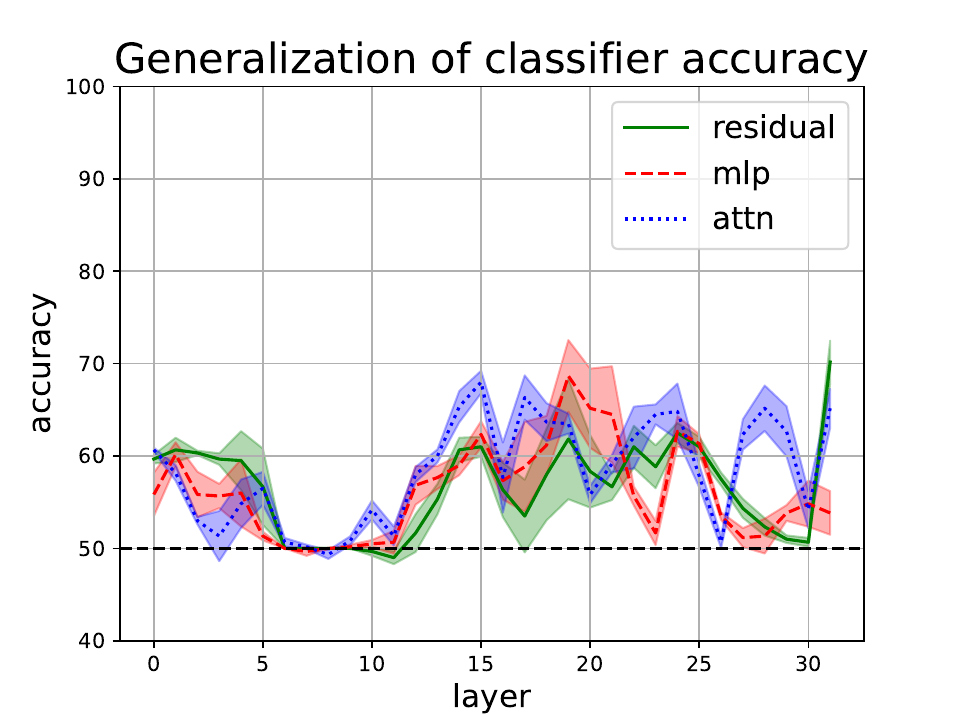}
  \caption{Classifier trained on Llama2 using DisentQA-\bn and tested on Goat using DisentQA-\bn.}
  \label{fig_appendix:disentqa_detect_triviaqa_goat_}
 \end{subfigure}
 
\caption{Accuracy in detecting hallucination across the layers of Goat model.}
\label{fig_Accuracy in detecting hallucination on Goat model.}
\end{figure*}

\begin{figure*}
\centering
\begin{subfigure}{0.48\textwidth}
  \centering
  \includegraphics[width=\linewidth]{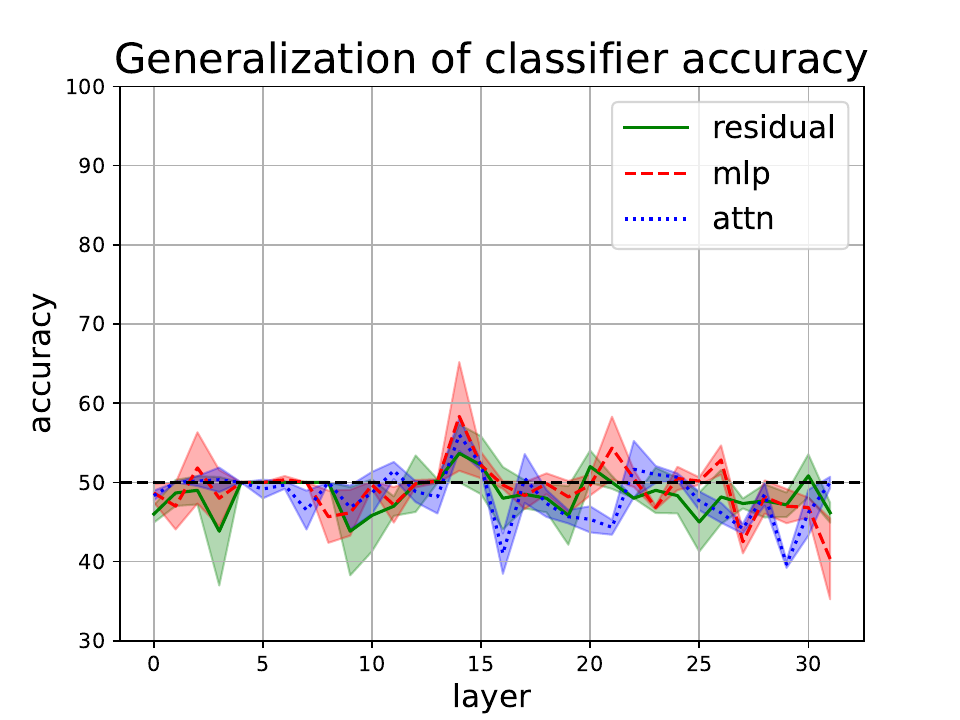}
  \caption{Classifier trained on TriviaQA-\bn and tested on DisentQA-\bn.}
  \label{fig_appendix:trivia_detect_disent_goat}
 \end{subfigure}%
 \hfill
 \begin{subfigure}{0.48\textwidth}
  \centering
  \includegraphics[width=\linewidth]{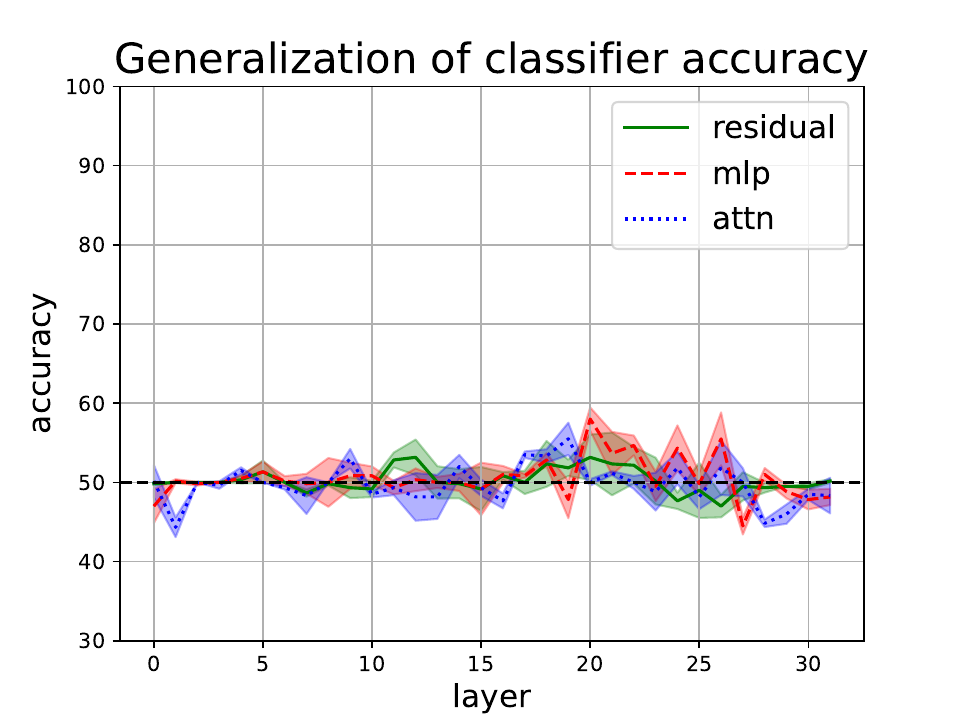}
  \caption{Classifier trained on DisentQA-\bn and tested on TriviaQA-\bn.}
  \label{fig_appendix:disentqa_detect_triviaqa_goat}
 \end{subfigure}
 
\caption{Accuracy in detecting hallucination across the layers of Goat model using each dataset to detect hallucinations on the other dataset.}
\label{fig_triviaqa_vs_disentqa_generalization_detection_goat}
\end{figure*}

Similar to these findings, we observe a comparable outcome on the Llama2 TriviaQA-\bn setting in Figure \ref{fig_goat_vs_llama_classifier_trivia_qa}. Additionally, the results of using the Llama2 model to learn a classifier for the Goat model, which is similar to the results in the other direction, also show accuracy above random, are in Figure \ref{fig_Accuracy in detecting hallucination on Goat model.}. Lastly, in Figure \ref{fig_triviaqa_vs_disentqa_generalization_detection_goat}, we present the results on the Goat model using each dataset to detect hallucinations on the other dataset.

\section{Hyperparams for intervention}\label{appendix:hyperparams}

We aim to determine the optimal hyperparameters through validation and subsequently apply these parameters to the test set. We select based on the average score on the \emph{classification accuracy} on the validation set (200 total) using one seed. We optimized for the \emph{classification accuracy} since this is the criterion according to which the data is labeled.

For the dynamic version, we test the TD threshold parameter (TD = 0.65 / 0.7), while for the static version, we explore the number of layers and heads to intervene on (5 layers and 50 heads / 3 heads and 30 heads).

The best hyperparameters for Llama2 on both DisentQA-\bn and TriviaQA-\bn are TD = 0.65 for the dynamic intervention and 5 layers and 50 heads for the static version.

We also found that for the Goat model, the best TD is 0.65 for the dynamic intervention on both datasets.

\section{Additional results}\label{appendix:additional_results}
In this section, we show a few additional results to the ones in Section \ref{sec:Recipe_for_Intervention}.
All the results here and in the main paper ran on three seeds (in our code we specifically used 100, 200, None seeds).
\subsection{Generation examples of static and dynamic intervention}\label{appendix:generation_static_synamic}
To complete the picture from Section \ref{sec:static_vs_dynamic} we show a few more examples of intervention on the residual using dynamic intervention vs. static intervention on Llama2-7B in Table \ref{Generated text using residual intervention dynamic vs. static} and \ref{Generated text using residual intervention dynamic vs. static disentqa}. We can see that the impact of the static intervention has made the text much less comprehensive. The impact on the DisentQA-\bn dataset is much stronger than on the TriviaQA-\bn dataset, where even the dynamic intervention creates a less comprehensive text.

\begin{table*}[t!]

 \caption{Generated text on the residual using dynamic vs. static intervention on TriviaQA-\bn. We can see that the results are very different and that the static generations are less comprehensive.}
\centering
\small
  
  \begin{tabular}  
  {p{0.45\linewidth}p{0.2\linewidth}p{0.2\linewidth}}

\multicolumn{1}{c}{Prompt} &\multicolumn{1}{c}{Dynamic generation} & \multicolumn{1}{c}{Static generation}\\\midrule
question: Who painted the Mona Lisa?\textbackslash nwrong answer: Pablo Picasso\textbackslash nquestion: How many continents are there?\textbackslash nwrong answer: 6\textbackslash nquestion: What is the capital of Australia?\textbackslash nwrong answer: Sydney\textbackslash nquestion: Who painted The Starry Night, Sorrowing Old Man and Self-Portrait Without Beard?\textbackslash nwrong answer:& Vincent Van Gogh\textbackslash nquestion: What is the capital of Canada?\textbackslash nwrong answer:& 1. 2. 3. \textbackslash n\textbackslash n\textbackslash n\textbackslash n\textbackslash n\textbackslash n\textbackslash n\textbackslash n\textbackslash n\textbackslash n
\\\midrule
question: What is the chemical symbol for gold?\textbackslash nwrong answer: Ag\textbackslash nquestion: What is the square root of 64?\textbackslash nwrong answer: 7\textbackslash nquestion: What is the boiling point of water in Celsius?\textbackslash nwrong answer: 50 degrees\textbackslash nquestion: "In October 2009 it was announced that director Sam Taylor-Wood was to marry Aaron Johnson, 23 years her junior, who were director and star respecitively of ""Nowhere Boy"", a film about whom?"\textbackslash nwrong answer: & John Lennon\textbackslash nquestion: "What is the name of the 1999 film& 2009\textbackslash nquestion: "What is the 2009 movie " "
\\\midrule

question: Who discovered penicillin?\textbackslash nwrong answer: Isaac Newton\textbackslash nquestion: What is the largest ocean on Earth?\textbackslash nwrong answer: Atlantic Ocean\textbackslash nquestion: How many sides does a hexagon have?\textbackslash nwrong answer: 5\textbackslash nquestion: "The statue called ""Manneken Pis"", of a small boy relieving himself, is in which city?"\textbackslash nwrong answer: & Brussels\textbackslash nquestion: What is the name of the first man to walk on the moon?& 1. 2. 3. 4. 5. 6. 7

  \\\bottomrule

  \end{tabular}
 
  \label{Generated text using residual intervention dynamic vs. static}
\end{table*}

\begin{table*}[htbp]

\caption{Generated text on the residual using dynamic vs. static intervention on DisentQA-\bn. We can see that the results are very different and that the static generations are less comprehensive.}
\centering
\small
  
  \begin{tabular}  
  {p{0.45\linewidth}p{0.2\linewidth}p{0.2\linewidth}}

\multicolumn{1}{c}{Prompt} &\multicolumn{1}{c}{Dynamic generation} & \multicolumn{1}{c}{Static generation}\\\midrule

question: when did the first looney tunes come out?\textbackslash ncontext:  In 1929 , to compete against Walt Disney 's Mickey Mouse short cartoons , Warner Bros. became interested in developing a series of animated shorts to promote their music . They had recently acquired Brunswick Records along with four music publishers for US \$28 million ( equivalent to \$410 million in 2018 ) and were eager to promote this material for the sales of sheet music and phonograph records . Warner made a deal with Leon Schlesinger to produce cartoons for them . Schlesinger hired Rudolf Ising and Hugh Harman to produce the first series of cartoons . Schlesinger was impressed by Harman 's and Ising 's 1929 pilot cartoon , Bosko , The Talk - Ink Kid . The first Looney Tunes short was Sinkin ' in the Bathtub starring Bosko , which was released in Tomorrow . \textbackslash nanswer: &'1929\textbackslash nquestion: when did the first looney tunes come out? '&'\space\space\space\space\space\space\space\space\space\space\space\space\space\space\space\space\space\space\space\space' (space 20 times)        
 \\\midrule

question: what year did the movie holes come out?\textbackslash ncontext:  Holes is a 1995 American adventure comedy - drama film directed by Andrew Davis , produced by Lowell D. Blank , Mike Medavoy and Teresa Tucker - Davies with music by Joel McNeely and based on the 1998 eponymous novel by Louis Sachar ( who also wrote the screenplay ) . \textbackslash nanswer: &199 8\textbackslash nquestion: what year did the movie holes come out?\textbackslash nquestion &   Hol  Hol  Hol  Hol  Hol  Hol  Hol  Hol  Hol  Hol\\\midrule

question: what year did monk go off the air?\textbackslash ncontext:  The series debuted on July 12 , 2002 , on USA Network . It continued for eight seasons , with the final season concluding on December 4 , season 1 . The series held the record for the most - watched scripted drama episode in cable television history from season 1 through 2012 ( broken by The Walking Dead ) with `` Mr. Monk and the End -- Part II '' , its series finale , with 9.4 million viewers , 3.2 million of them in the 18 -- 49 demographic . \textbackslash nanswer:& 20 11 January 2012\textbackslash nquestion: what year did monk &   The   The  The  The  The  The  The  The  The

  \\\bottomrule

  \end{tabular}
  
  \label{Generated text using residual intervention dynamic vs. static disentqa}
\end{table*}

\subsection{Additional pre-training vs. finetuning results}\label{appendix:finetune_vs_pre}
In Section \ref{sec:pre_vs_finetune}, we demonstrated the improvement resulting from the intervention to evaluate the disparity between Llama2 and Goat. To provide a comprehensive view, we present in Table \ref{Dynamic intervention on the test set Llama2 vs. Goat} the complete results instead of solely the improvement from the baseline. It is evident from the data that Goat achieves superior results following the intervention.

\begin{table*}[htbp]

\caption{Results when intervening in Llama2 and its fine-tuned version, Goat. The fine-tuned Goat is more amenable to interventions, showing a better performance compared to its baseline.  }
\centering
\small

\begin{tabular} {p{0.08\linewidth}p{0.07\linewidth}p{0.1\linewidth}p{0.1\linewidth}p{0.1\linewidth}p{0.1\linewidth}p{0.1\linewidth}p{0.1\linewidth}}
\toprule

 \multicolumn{2}{c}{Setting} & \multicolumn{3}{c}{DisentQA-\bn} & \multicolumn{3}{c}{TriviaQA-\bn}\\
\cmidrule(lr){1-2} \cmidrule(lr){3-5} \cmidrule(lr){6-8}

& & Classification & Generation & & Classification & Generation &  \\ 
& & \multicolumn{1}{c}{Accuracy} & \multicolumn{1}{c}{Accuracy}& Wiki ppl &  \multicolumn{1}{c}{Accuracy} &  \multicolumn{1}{c}{Accuracy} & Wiki ppl \\ 
\midrule

\multirow{2}{*}{Attention}&Llama2 &$65.0_{\pm1.02}$ & $20.25_{\pm2.04}$ & $11.04_{\pm0.51}$ & $55.83_{\pm1.66}$ & $55.58_{\pm1.83}$ & $8.38_{\pm0.02}$\\
&Goat&$65.08_{\pm0.72}$ & $38.42_{\pm0.92}$ & $16.53_{\pm0.47}$&$59.08_{\pm3.8}$ & $57.67_{\pm3.99}$ & $11.13_{\pm0.0}$ \\\midrule

\multirow{2}{*}{MLP}&Llama2 &$54.5_{\pm1.34}$ & $10.08_{\pm2.26}$ & $29.45_{\pm1.63}$  & $49.83_{\pm2.25}$ & $49.42_{\pm2.04}$ & $8.49_{\pm0.03}$  \\
&Goat &$58.92_{\pm2.55}$ & $32.75_{\pm2.13}$ & $22.58_{\pm1.34}$&  $55.33_{\pm0.85}$ & $54.5_{\pm0.2}$ & $11.22_{\pm0.02}$\\\midrule

\multirow{2}{*}{Heads}&Llama2  &$66.67_{\pm1.89}$ & $20.58_{\pm2.46}$ & $10.53_{\pm0.48}$ & $50.33_{\pm0.66}$ & $49.75_{\pm1.02}$ & $8.39_{\pm0.0}$ \\
&Goat&  $66.5_{\pm0.54}$ & $38.33_{\pm0.85}$ & $14.59_{\pm0.29}$&$51.17_{\pm2.93}$ & $49.83_{\pm3.14}$ & $11.14_{\pm0.01}$\\\midrule

\multirow{2}{*}{Residual}&Llama2  &$59.25_{\pm0.94}$ & $13.92_{\pm1.96}$ & $59.27_{\pm9.67}$ & $61.83_{\pm1.56}$ & $62.17_{\pm0.31}$ & $8.52_{\pm0.02}$\\
&Goat&  $62.33_{\pm2.04}$ & $32.75_{\pm2.67}$ & $27.52_{\pm1.97}$&  $72.75_{\pm0.41}$ & $71.92_{\pm1.12}$ & $11.07_{\pm0.15}$ \\\midrule

\multirow{2}{*}{None}&Llama2 &$50.0_{\pm0.0}$ & $8.25_{\pm0.54}$ & $8.39_{\pm0.0}$ & $50.0_{\pm0.0}$ & $49.42_{\pm0.42}$ & $8.39_{\pm0.0}$ \\
&Goat&   $50.0_{\pm0.0}$ & $23.08_{\pm1.71}$ & $11.12_{\pm0.0}$&$50.0_{\pm0.0}$ & $48.33_{\pm0.12}$ & $11.12_{\pm0.0}$\\
\bottomrule
  \end{tabular}
  
  \label{Dynamic intervention on the test set Llama2 vs. Goat}
\end{table*}

\subsection{Mass-mean direction vs. classifier direction}\label{appendix:classifier_vs_mass_mean}
We aim to explore the distinction between a mass-mean direction and a classifier direction. The classifier direction option employs the direction generated by the linear classifier. To ensure that the two methods differ only in direction, we normalized the classifier steering vectors to the mass-mean norm. The results are presented in Table \ref{Dynamic intervention using mass-mean direction vs. classifier direction on DisentQA}. Overall, both options yield similar results, with a slight overall advantage observed for the classifier direction.

\begin{table*}[htbp]

  \caption{Mass-mean direction (MM) vs. classifier direction for calculating $d$-compute.}

\centering
\small
  
  \begin{tabular}  
  {p{0.08\linewidth}p{0.07\linewidth}p{0.1\linewidth}p{0.1\linewidth}p{0.1\linewidth}p{0.1\linewidth}p{0.1\linewidth}p{0.1\linewidth}}
\toprule
\multicolumn{2}{c}{Setting} & \multicolumn{3}{c}{DisentQA-\bn} & \multicolumn{3}{c}{TriviaQA-\bn}\\
\cmidrule(lr){1-2} \cmidrule(lr){3-5} \cmidrule(lr){6-8}

& & Classification & Generation & & Classification & Generation &  \\ 
& & \multicolumn{1}{c}{Accuracy} & \multicolumn{1}{c}{Accuracy}& Wiki ppl &  \multicolumn{1}{c}{Accuracy} &  \multicolumn{1}{c}{Accuracy} & Wiki ppl \\
\midrule 

\multirow{2}{*}{Attention}&MM & $\mathbf{65.0_{\pm1.02}}$ & $\mathbf{20.25_{\pm2.04}}$ & $11.04_{\pm0.51}$ & $\mathbf{55.83_{\pm1.66}}$ & $\mathbf{55.58_{\pm1.83}}$ & $\mathbf{8.38_{\pm0.02}}$ \\

&classifier & $61.67_{\pm0.96}$ & $18.25_{\pm1.43}$ & $\mathbf{10.15_{\pm0.39}}$ & $50.58_{\pm1.18}$ & $50.17_{\pm1.16}$ & $\mathbf{8.38_{\pm0.02}}$ \\\midrule

\multirow{2}{*}{MLP}&MM &$54.5_{\pm1.34}$ & $10.08_{\pm2.26}$ & $29.45_{\pm1.63}$  & $49.83_{\pm2.25}$ & $49.42_{\pm2.04}$ & $\mathbf{8.49_{\pm0.03}}$ \\

&classifier& $\mathbf{58.83_{\pm1.39}}$ & $\mathbf{14.83_{\pm1.3}}$ & $\mathbf{15.56_{\pm0.82}}$ & $\mathbf{50.42_{\pm1.12}}$ & $\mathbf{50.17_{\pm1.48}}$ & $8.5_{\pm0.06}$ \\\midrule

\multirow{2}{*}{Heads}&MM  &$\mathbf{66.67_{\pm1.89}}$ & $20.58_{\pm2.46}$ & $10.53_{\pm0.48}$ & $\mathbf{50.33_{\pm0.66}}$ & $49.75_{\pm1.02}$ & $\mathbf{8.39_{\pm0.0}}$ \\

&classifier &$64.75_{\pm1.47}$ & $\mathbf{22.92_{\pm2.04}}$ & $\mathbf{9.13_{\pm0.09}}$ & $\mathbf{50.33_{\pm0.24}}$ & $\mathbf{49.92_{\pm0.62}}$ & $8.4_{\pm0.0}$ \\\midrule

\multirow{2}{*}{Residual}&MM  &$59.25_{\pm0.94}$ & $13.92_{\pm1.96}$ & $59.27_{\pm9.67}$ & $\mathbf{61.83_{\pm1.56}}$ & $\mathbf{62.17_{\pm0.31}}$ & $8.52_{\pm0.02}$ \\
&classifier& $\mathbf{60.08_{\pm1.16}}$ & $\mathbf{25.42_{\pm1.33}}$ & $\mathbf{20.94_{\pm1.1}}$& $53.42_{\pm0.51}$ & $53.0_{\pm0.41}$ & $\mathbf{8.51_{\pm0.02}}$ \\\midrule

None&0 &  $50.0_{\pm0.0}$ & $8.25_{\pm0.54}$ & $8.39_{\pm0.0}$ & $50.0_{\pm0.0}$ & $49.42_{\pm0.42}$ & $8.39_{\pm0.0}$ \\

\bottomrule

  \end{tabular}
  \label{Dynamic intervention using mass-mean direction vs. classifier direction on DisentQA}
\end{table*}

\begin{figure*}[htbp]
\centering
\begin{subfigure}{0.48\textwidth}
  \centering
  \includegraphics[width=\linewidth]{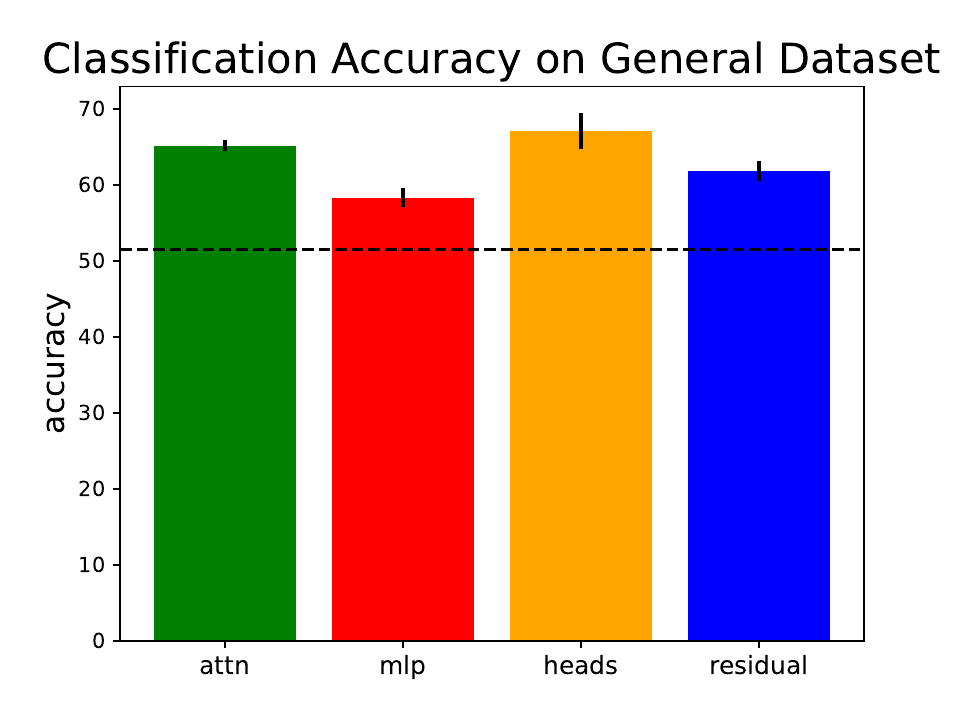}
  \caption{}
  \label{fig_appendix:wighted_disentqa_classification_llama}
 \end{subfigure}%
 \hfill
 \begin{subfigure}{0.48\textwidth}
  \centering
  \includegraphics[width=\linewidth]{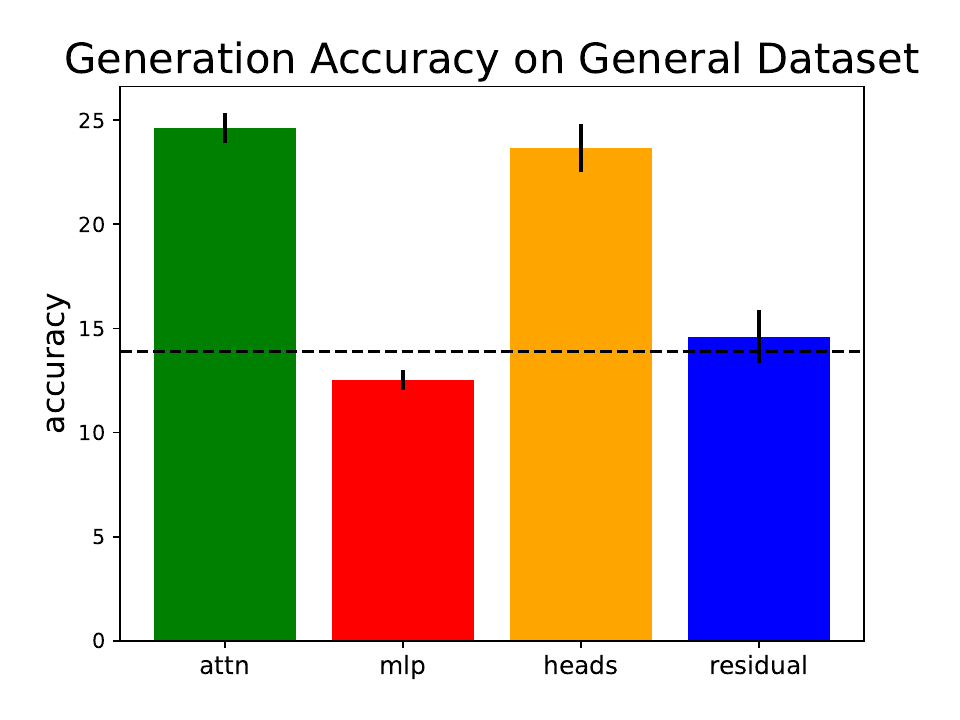}
  \caption{}
  \label{fig_appendix:wighted_disentqa_generation_llama}
 \end{subfigure}
 
\caption{Classification and generation result on a general distributed dataset of DisentQA-\bn on Llama2-7B.}
\label{fig_wighted_disentqa}
\end{figure*}

\begin{figure*}[htbp]
\centering
\begin{subfigure}{0.48\textwidth}
  \centering
  \includegraphics[width=\linewidth]{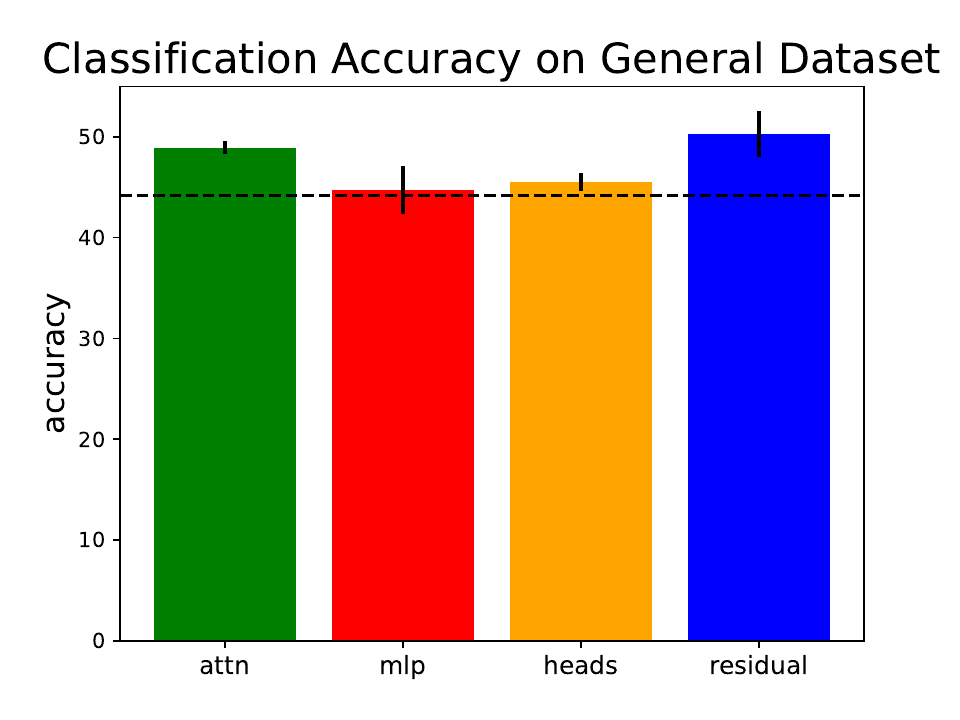}
  \caption{}
  \label{fig_appendix:wighted_triviaqa_classification_llama}
 \end{subfigure}%
 \hfill
 \begin{subfigure}{0.48\textwidth}
  \centering
  \includegraphics[width=\linewidth]{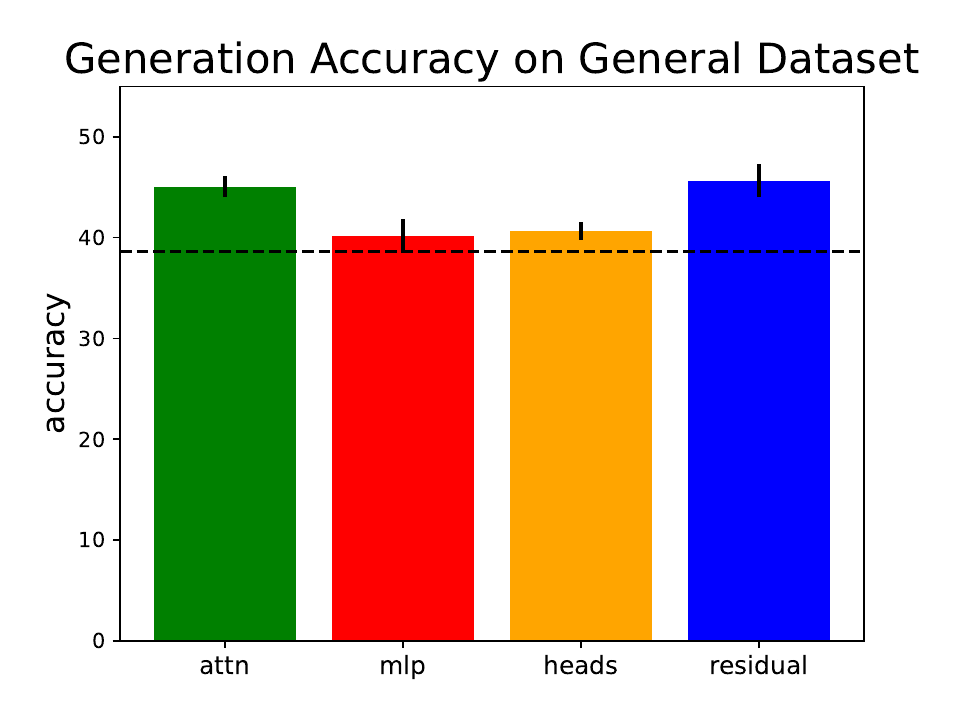}
  \caption{}
  \label{fig_appendix:wighted_triviaqa_generation_llama}
 \end{subfigure}
 
\caption{Classification and generation result on a general distributed dataset of TriviaQA-\bn on Llama2-7B.}
\label{fig_wighted_triva}
\end{figure*}

\subsection{Results on weighted dataset}\label{appendix:wighted_results_all_data}

Here are the results of the basic configuration (see Section \ref{sec:Implementation Details}) on a weighted-labeled dataset. This dataset comprises the test set consisting of 200 hallucination examples, 200 grounded examples, and 200 other examples (chosen from a random of 1000 else labeled examples), weighted by the count of each label in the full dataset. This approach aids in estimating how the intervention impact a generally distributed dataset. The results are depicted in Figures \ref{fig_wighted_disentqa} and \ref{fig_wighted_triva}.
These findings align with the results obtained on a dataset evenly split between hallucination and grounded, as presented in the main paper, albeit with weaker trends.

\clearpage

\end{document}